\documentclass[letterpaper]{article} % DO NOT CHANGE THIS
\usepackage{aaai23}

\usepackage{times}  % DO NOT CHANGE THIS
\usepackage{helvet}  % DO NOT CHANGE THIS
\usepackage{courier}  % DO NOT CHANGE THIS
\usepackage[hyphens]{url}  % DO NOT CHANGE THIS
\usepackage{graphicx} % DO NOT CHANGE THIS
\usepackage{natbib}  % DO NOT CHANGE THIS AND DO NOT ADD ANY OPTIONS TO IT
\usepackage{caption} % DO NOT CHANGE THIS AND DO NOT ADD ANY OPTIONS TO IT
\usepackage{algorithm}
\usepackage{algorithmic}
\usepackage{mathtools}
\usepackage{cleveref}

\usepackage{multirow}
\usepackage{hhline}
\usepackage{lipsum}
\usepackage{makecell}
\usepackage{xcolor} 
\usepackage{microtype} 
\usepackage{booktabs}
\usepackage{amsfonts}
\usepackage{nicefrac}
\usepackage{subcaption}
\usepackage{newfloat}
\usepackage{listings}
\DeclareCaptionStyle{ruled}{labelfont=normalfont,labelsep=colon,strut=off} % DO NOT CHANGE THIS
\floatstyle{ruled}
\newfloat{listing}{tb}{lst}{}
\floatname{listing}{Listing}
\newcommand{\ourmodel}{GeoDTR}
\newcommand{\Rone}{R@$1$}
\newcommand{\Rfive}{R@$5$}
\newcommand{\Rten}{R@$10$}
\newcommand{\Ronep}{R@$1\%$}
\newcommand{\aug}{LS}

\newcommand{\B}[1]{\textcolor{magenta}{\textbf{#1}}}
\newcommand{\SB}[1]{\textcolor{cyan}{\textbf{#1}}}

\newcommand{\FT}[1]{{\textbf{#1}}}
\newcommand{\xh}[1]{\textcolor{black}{#1}}

\title{Cross-view Geo-localization via Learning Disentangled Geometric Layout  Correspondence}
\author{
    Xiaohan Zhang\textsuperscript{\rm 1,2}\equalcontrib, Xingyu Li\textsuperscript{\rm 3}\equalcontrib, Waqas Sultani\textsuperscript{\rm 4}, Yi Zhou\textsuperscript{\rm 5}, Safwan Wshah\textsuperscript{\rm 1,2}\thanks{Corresponding and senior author.}
}
\affiliations{
    \textsuperscript{\rm 1}Department of Computer Science, University of Vermont, Burlington, USA\\
    
    \textsuperscript{\rm 2}Vermont Complex Systems Center, University of Vermont, Burlington, USA\\

    \textsuperscript{\rm 3} Shanghai Center for Brain Science and Brain-Inspired Technology, China\\
    \textsuperscript{\rm 4} Intelligent Machine Lab, Information Technology University, Pakistan\\
    \textsuperscript{\rm 5} NEL-BITA, School of Information Science and Technology, University of Science and Technology of China, China \\
}

\usepackage{bibentry}
\begin{document}

\maketitle

\begin{abstract}
Cross-view geo-localization aims to estimate the location of a query ground image by matching it to a reference geo-tagged aerial images database. As an extremely challenging task, its difficulties root in the drastic view changes and different capturing time between two views. Despite these difficulties, recent works achieve outstanding progress on cross-view geo-localization benchmarks. However, existing methods still suffer from poor performance on the cross-area benchmarks, in which the training and testing data are captured from two different regions. We attribute this deficiency to the lack of ability to extract the spatial configuration of visual feature layouts and models' overfitting on low-level details from the training set. In this paper, we propose \ourmodel{} which explicitly disentangles geometric information from raw features and learns the spatial correlations among visual features from aerial and ground pairs with a novel geometric layout extractor module. This module generates a set of geometric layout descriptors, modulating the raw features and producing high-quality latent representations. In addition, we elaborate on two categories of data augmentations, (i) Layout simulation, which varies the spatial configuration while keeping the low-level details intact. (ii) Semantic augmentation, which alters the low-level details and encourages the model to capture spatial configurations. These augmentations help to improve the performance of the cross-view geo-localization models, especially on the cross-area benchmarks. Moreover, we propose a counterfactual-based learning process to benefit the geometric layout extractor in exploring spatial information. Extensive experiments show that \ourmodel{} not only achieves state-of-the-art results but also significantly boosts the performance on \xh{same-area and cross-area benchmarks.} Our code can be found at \textcolor{magenta}{\url{https://gitlab.com/vail-uvm/geodtr}}.
\end{abstract}

\section{Introduction}
\label{sec::introduction}

Cross-view geo-localization is defined as the estimation of the location of a ground image (also known as query image) from a set of geo-tagged aerial images (also known as reference images). \xh{The ground images are usually captured from cameras mounted on vehicles or taken by pedestrians.} Cross-view geo-localization can be applied in different fields such as autonomous driving~\citep{kim2017satellite}, unmanned aerial vehicle navigation~\citep{UAV1}, and augmented reality~\citep{AR1}. Most of the existing cross-view geo-localization methods~\citep{SAFA,DSM,l2ltr,cvmnet,Vo,CVUSA,comingDown,featureTransport,liu2019lending,wang2021each,zheng2020university} frame the problem as a retrieval task. These methods normally train a model to push the \xh{corresponding aerial image and ground image pairs (also known as aerial-ground pairs)} closer in latent space and push the unmatched pairs further away from each other. At the deployment, the location of an aerial image with the highest similarity is the prediction for a given query ground image.

Cross-view geo-localization is considered an extremely challenging problem because of 1) the drastic change of viewpoints, 2) the difference in capturing time, 3) and the different resolutions between ground and aerial images~\xh{\cite{survey}}. Tackling these challenges requires a detailed understanding of the image content and the spatial configuration of visual features, e.g., buildings and roads. Most existing methods~\citep{SAFA, featureTransport,cvmnet,lin2015learning,hardTriplet,Vo} match ground and aerial images by exploiting features extracted from convolutional neural networks (CNNs). For instance, \citet{SAFA} directly encodes the relative positions among object features by using the Spatial-aware Position Embedding (SPE) module. These methods are limited in exploring the spatial configuration of visual features which is a global property. Recently, with the advancement of Transformer~\citep{attention}, several methods~\citep{l2ltr, transgeo} explore extracting latent features from global contextual information. However, such methods solely rely on multi-head attention mechanism to implicitly explore correlations in the input features. Consequently, the correlations are unavoidably entangled in those approaches. 
% \xh{However, such methods rely on large pretrained ViT~\citep{ViT} models that need large amount of GPU memory~\citep{l2ltr} or a 2-stage training paradigm with complicated optimizer~\citep{transgeo}.}

This paper introduces \textit{\ourmodel{}} which processes the low-level feature and spatial configuration of visual features separately. 
To capture spatial configuration, we propose a novel geometric layout extractor sub-module.
% What lies at the heart of model is the geometric layout extractor module. 
This sub-module generates a set of geometric layout descriptors that reflects the global contextual information among visual features in an image.
% These descriptors turns out to be critical for geo-localization performance. 
% The correlations between visual features are explored by the geometric layout extractor which is a transformer-based sub-module in our model. 
% To further strengthen the quality of those learned geometric layout descriptors, we integrate two novel techniques 
We strengthen the quality of the geometric layout descriptors through Layout simulation, Semantic augmentation (LS) and \xh{a} counterfactual (CF) training schema. LS generates different layouts for aerial and ground pairs with perturbed low-level details which improves the diversity of training aerial-ground pairs. Unlike existing data augmentation methods in cross-view geo-localization, LS \textit{maintains the geometric/spatial correspondence during training}. Thus, it can be universally applied to any geo-localization method. Moreover, we observe that LS improves the performance on cross-area experiments because of the regularization of LS. We introduce a novel distance-based counterfactual (CF) training schema to fortify the learning of the extracted descriptors. Specifically, it provides auxiliary supervision to the geometric layout extractor to refine global contextual information.
%Different from the existing approaches, \ourmodel{} extracts the spatial configuration without engaging low-level details. Compared with ViT based methods, our model is much smaller.
% Different from recent transformer-based models, \ourmodel{} extracts the spatial configuration without engaging low-level details and therefore results in an efficient model. 
The performance of our proposed model,
\ourmodel{}, shows a substantial increase and achieves state-of-the-art compared to other algorithms on the common cross-view geolocalization datasets, CVUSA~\citep{CVUSA} and CVACT~\citep{liu2019lending}. 
% In practice, machine learning models are expected not only to be deployed where the training data was collected.
% Current cross-view geo-localization approaches struggle in generalizing to other unseen areas~\citep{l2ltr} as reported in the current state-of-the-art methods. We believe this is due to the use of siamese networks which can easily take a shortcut by exploiting low-level details. To tackle this issue, we propose a layout simulation technique and semantic augmentation to ameliorate models from overfitting to low-level details.
% The layout simulation generates virtual aerial and ground image pairs with realistic spatial configurations. Semantic augmentation perturbs the low-level details while maintaining the spatial configurations.
Our contributions can be summarized as threefold:

\begin{itemize}
    \item We propose \ourmodel{} which disentangles geometric information from raw features to increase the transformer efficiency. The proposed model effectively explores the spatial configurations and low-level details, and better captures the correspondence between aerial and ground images.
    %The proposed model effectively capture both spatial configurations and low-level details between aerial and ground images. 
    %The captured spatial configuration is shown to be helpful in producing high-quality latent representations.
    %which is specialized in capturing geometric layout from both aerial images and street images.
    %The captured geometric correspondence is proved to be useful in generating latent features for cross-view geo-localization. 
    
    %The proposed \ourmodel{} achieves state-of-the-art results on standard, fine-grained, and cross-area cross-view geo-localization benchmarks.
    
    \item We propose layout simulation and semantic augmentation techniques that improve the performance of \ourmodel{} (as well as existing methods) on cross-area experiments. %As illustrated by our experiments, these techniques also improve existing cross-view geo-localization methods.
    
    %Our extensive experiments demonstrate that the layout simulation benefits the \ourmodel{} in capturing the correlation among objects and the semantic augmentation boosts the performance of \ourmodel{}.
    
    \item We introduce a novel counterfactual-based learning schema that guides \ourmodel{} to better grasp the spatial configurations and therefore produce better latent feature representations.
    
    %Our experiments show that this learning progress improves the performance of \ourmodel{} in different benchmarks.
\end{itemize}

\section{Related works}
\label{sec::related_works}
\subsection{Cross-view Geo-localization}
% \noindent\textbf{Feature-based cross-view geo-localization:}
\subsubsection{Feature-based Cross-view Geo-localization}
Feature-based geo-localization methods extract both aerial and ground latent representations from local information using CNNs~\citep{lin2013cross, lin2015learning, CVUSA}. \xh{Existing works studied different aggregation strategy~\citep{cvmnet}, training paradigm~\citep{Vo}, loss functions (i.e. HER~\citep{hardTriplet}) and SEH~\citep{SEH}} \xh{and feature transformation (i.e. feature fusion~\citep{bridging} and Cross-View Feature Transport (CVFT)~\citep{featureTransport}).}
The above-mentioned feature-based methods did not fully explore the effectiveness of spatial information due to the locality of CNN which lacks of ability to explore global correlations. By leveraging the ability to capture global contextual information of the transformer, our \ourmodel{} learns the geometric correspondence between ground images and aerial images through a transformer-based sub-module which results in a better performance.

% \noindent\textbf{Geometry-based cross-view geo-localization:}
\subsubsection{Geometry-based Cross-view Geo-localization}
% Besides the feature-based methods, many studies also investigate the geometric and contextual correspondence between aerial and street views.
Recently, learning to match the geometric correspondence between aerial and street views is becoming a hot topic. \citet{liu2019lending} proposed to train the model with encoded camera orientation in aerial and ground images.
\citet{SAFA} proposed SAFA which aggregates features through its learned geometric correspondence from ground images and polar transformed aerial images. Later, the same author proposed Dynamic Similarity Matching (DSM)~\citep{DSM} to geo-localizing limited field-of-view ground images by a sliding-window-like algorithm. \xh{CDE~\citep{comingDown}} combined GAN~\citep{gan} and SAFA~\citep{SAFA} to learn cross-view geo-localization and ground image generation simultaneously. Despite the remarkable performance achieved by these geometric-based methods, they are limited by the nature of CNNs which explores the local correlation among pixels. On the other hand, \ourmodel{} not only explicitly models the local correlation but also explores the global contextual information through a transformer-based sub-module. 
The quality of this global contextual information is further strengthened by our CF learning schema and \aug{} technique. Finally, benefitted from the model design, GeoDTR does not solely rely on polar transformed aerial view.

Recent researches~\citep{l2ltr,transgeo} also explore to capture non-local correlations in the images. L2LTR~\citep{l2ltr} studied a hybrid ViT-based~\citep{ViT} methods while TransGeo~\citep{transgeo} proposed a pure transformer-based model. The above mentioned methods implicitly model the spatial information from the raw features because of solely rely on transformer. Nevertheless, our \ourmodel{} explicitly disentangles low-level details and spatial information from raw features. Moreover, \ourmodel{} has less trainable parameter than L2LTR~\citep{l2ltr} and does not require the 2-stage training paradigm as proposed in TransGeo~\citep{transgeo}.

% \noindent\textbf{Data augmentation in cross-view geo-localization:}
\subsubsection{Data Augmentation in Cross-view Geo-localization}
% Even though data augmentation is popular in computer vision, it is seldom explored in cross-view geo-localization. %Applying traditional data augmentation such as random crop to cross-view geo-localization undermines the spatial layout in aerial and ground image pairs, making these methods less applicable in this area.
Data augmentation is popular in computer vision. Nonetheless, data augmentation in cross-view geo-localization is very limited due to the vulnerability of the spatial correspondence between aerial and ground images which can be easily sabotaged by a minor interference. For instance, most existing methods~\citep{liu2019lending, Rodrigues_2022_WACV, Vo, hardTriplet} randomly rotate or shift one view while fixing the other one. On the other hand, \citet{rodrigues2021these} randomly blackout ground objects according to their segmentation from street images. In this paper, we propose LS techniques that \textit{maintain} geometric correspondence between images of the two views while varying geometric layout and visual features during the training phase. Our extensive experiments demonstrate that LS can significantly improve the performance on cross-area datasets not only for \ourmodel{} but also can be universally applied to other existing methods.
% Recently, \citet{liu2019lending} proposed to randomly shift the input ground-level image while fixing the corresponding aerial image. \citet{rodrigues2021these} proposed a data augmentation pipeline that utilizes semantic segmentation maps of ground images to randomly erase some objects. Similar to~\citep{liu2019lending}, \citet{Rodrigues_2022_WACV} proposed to randomly rotate aerial images while fixing ground images to augment data.
% Data augmentations proposed in~\citep{liu2019lending,Rodrigues_2022_WACV,rodrigues2021these} undermine the geometric correspondence of visual features in aerial images and ground images. In this paper, we propose data augmentation techniques that maintain geometric correspondence while varying visual features.
%In this paper, our proposed layout simulation varies geometric layout while keeps the geometric correspondence. The semantic augmentation maintains the same geometric layout while varies the low-level details.

\subsection{Counterfactual Learning}
%Causal inference is emerging in recent years. 
The idea of counterfactual in causal inference~\citep{Pearl_Causality} has been successfully applied in several research areas such as explainable artificial intelligence ~\citep{explain_cf}, visual question answering~\citep{VQA_cf}, physics simulation~\citep{cophy_cf}, and reinforcement learning~\citep{reinforce_cf}. In this work, we propose a novel distance-based counterfactual (CF) learning schema which strengthens the quality of learned geometric descriptors for our \ourmodel{}. Experiments show that the proposed CF learning schema improves the performance of \ourmodel{}.

% \citet{l2ltr} proposed L2LTR which was built upon a pretrained Vision Transformer~\citep{ViT} to capture the global contextual information.} \xh{More recently, \citep{transgeo} proposed TransGeo which is a 2-stage trained transformer based cross-view geo-localization model. Distinct from TransGeo, our model does not require the global mining pool and advanced optimizer, for example, ASAM~\citep{ASAM}.}

% The main point of counterfactual inference is the causal reasoning between an observed event (factual) and an imaginary event (counterfactual).
% Existing works had applied counterfactual inference to model training in different research areas, for example, vision-language grounding~\citep{ground_cf}, visual question answering~\citep{VQA_cf}, physics simulation~\citep{cophy_cf}, and reinforcement learning~\citep{reinforce_cf}. 

\section{Methodology}
\label{sec::methodology}

\begin{figure*}[!t]
    \centering
    \includegraphics[page=5,width=\textwidth, clip,trim=0.5cm 69.5cm 0.5cm 1cm]{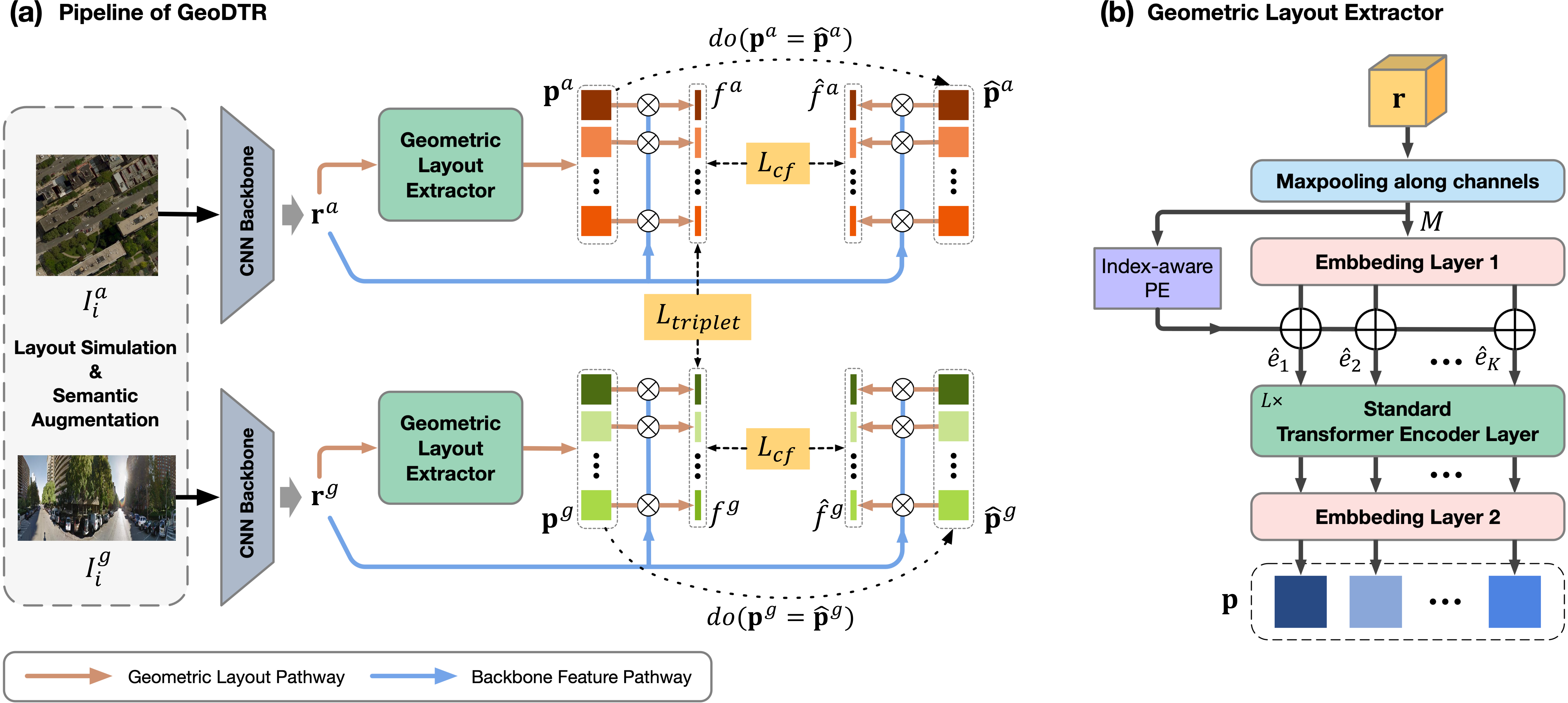}
    \caption{(a) The overview pipeline of our proposed model \ourmodel{}. (b) Illustration of our proposed geometric layout extractor.}
    \label{fig:model}
\end{figure*}

\subsection{Problem Formulation\label{sec:problemformulation}}

Considering a set of ground-aerial image pairs $\{(I^g_i, I^a_i)\}, i=1,\dots,N$, where superscripts $g$ and $a$ are abbreviations for ground and aerial, respectively, and $N$ is the number of pairs.
Each pair is tied to a distinct geo-location. 
% More specifically, we assume the camera position of a ground image coincides with the center of the corresponding aerial image.
In the cross-view geo-localization task, given a query ground image $I^g_q$ with index $q$, one searches for the best-matching reference aerial image $I^a_b$ with $b\in\{1,\dots,N\}$.

For the sake of a feasible comparison between a ground image and an aerial image, we seek discriminative latent representations $f^g$ and $f^a$ for the images.
% we seek functions $\phi^g(\cdot)$ and $\phi^a(\cdot)$ that map the images into proper latent representations. 
These representations are expected to 
capture the dramatic view-change as well as the abundant low-level details, such as textual patterns.
Then the image retrieval task can be made explicit as
\begin{equation}\label{eq:task}
    b = \underset{i\in\{1, \dots, N\}}{\arg\min}\, d(f^g_q, f^a_i),
\end{equation}
where $d(\cdot,\cdot)$ denotes the $L_2$ distance.
% and $f^v=\phi^v(I^v)$ are the latent representations. 
For the compactness in symbols, we will use superscript $v$ for cases that apply to both ground ($g$) and aerial ($a$) views. We adopt this convention throughout the paper.

\subsection{Geometric Layout Modulated Representations}
To generate high-quality latent representations for cross-view geo-localization, we emphasize the spatial configurations of visual features as well as low-level features. The spatial configuration reflects not only the positions but also the global contextual information among visual features in an image. 
One could expect such geometric information to be stable during the view-change.
% Specifically, spatial configuration
% of visual features 
% represents the positional correlation among visual features in the ground (aerial) view images
% between ground and aerial  visual features 
% and is of geometric nature. Hence, we referred to this spatial configuration as geometric layout.
Meanwhile, the low-level features such as color and texture, help to identify visual features across different views. 

Specifically, we propose the following decomposition of the latent representation
\begin{equation}
    f^v = \mathbf{p}^v \circ \mathbf{r}^v.
    \label{eq:main}
\end{equation}
$\mathbf{p}^v=\{p^v_{\xh{m}}\}_{{\xh{m}}=1,\dots,K}$ is the set of $K$ geometric layout descriptors that summarize the spatial configuration of visual features, 
% with $p^v_i \in\mathbb{R}^{H\times W}$,
and $\mathbf{r}^v=\{r^v_j\}_{j=1,\dots,C}$ \xh{denotes} the raw latent representations of $C$ channels that is generated by any backbone encoder. Both $p^v_{\xh{m}}$ and $r^v_j$ are vectors in $\mathbb{R}^{H\times W}$ with $H$ and $W$ being the height and width of the raw latent representations, respectively.
The modulation operation $\mathbf{p}^v \circ \mathbf{r}^v$ expands as 
\begin{equation}
\left( \langle p^v_1, r^v_1 \rangle,\dots, \langle p^v_1, r^v_C \rangle, \dots,  \langle p^v_K, r^v_1 \rangle,\dots, \langle p^v_K, r^v_C \rangle \right),
\end{equation}
where $\langle p^v_{\xh{m}}, r^v_j \rangle$ denotes the Frobenius inner product of $p^v_{\xh{m}}$ and $r^v_j$.
In this sense, the resulting $f^v \in \mathbb{R}^{CK}$ are referred to as the \emph{geometric layout modulated representations} and will be fed to~\Cref{eq:task} to retrieve the best-matching aerial images.
Our model design closely follows the above decomposition.
% with $r^v_j \in\mathbb{R}^{H\times W}$.

% Our \ourmodel{} explicitly processes these two distinct aspects. In the following, we overview our model and provide detailed descriptions for each sub-module therein.

%(i) Spatial configuration of visual features. This represents the positional correlation between ground and aerial  visual features and is of geometric nature. Hence, we referred this spatial configuration as geometric layout.
%(ii) The low-level features such as color and texture, which help to identify visual features across different views. Our \ourmodel{} explicitly process these two distinct aspects. In the following, we overview our model and provide detailed descriptions for each sub-module therein.
% \textcolor{red}{Do we also have specific name for it, If yes, please mention for consistancy with (i)}

%To address the above two aspects,

\subsection{\ourmodel{} Model}

\subsubsection{Model Overview}
\ourmodel{} (see~\Cref{fig:model} (a)) is a siamese neural network including two branches for the ground and the aerial views, respectively.
Within a branch, there are two distinct processing pathways, i.e., the backbone feature pathway and the geometric layout pathway.
In the backbone feature pathway, a CNN backbone encoder processes the input image to generate raw latent representations $\mathbf{r}^v$ \xh{where $v=g \text{ or } v=a$}. Due to the nature of the CNN backbone, these representations carry the positional information as well as the low-level feature information.

% \noindent\textbf{The geometric layout pathway:}
The geometric layout pathway is devoted to exploring the global contextual information among visual features. 
This pathway includes a core sub-module called the geometric layout extractor, 
which generates a set of geometric layout descriptors $\mathbf{p}^v$ based on the raw latent representations $\mathbf{r}^v$.
These descriptors will modulate $\mathbf{r}^v$, integrating the geometric layout information therein.
With a stand-alone treatment of the geometric layout, one avoids introducing undesired correlations among the low-level features from different visual features. In the following, we will describe the key components of \ourmodel{} in detail.

% Let $r^c \in\mathbb{R}^{H\times W}, c\in\{1,2,\dots,C\}$ and $p^k\in \mathbb{R}^{H\times W}, k\in\{1,2,\dots,K\}$ 
% Let $r^c, c\in\{1,2,\dots,C\}$ and $p^k, k\in\{1,2,\dots,K\}$ 
% be a channel-wise intermediate representation and the corresponding geometric layout descriptors, respectively. Here, $C$ is the number of channels, and $K$ is the number of descriptors.
% , and $H,W$ refer to the sizes of the intermediate representations.
% Then the $\circ$ operator in~\Cref{eq:main} expands as following
% \begin{equation}\label{eq:descriptor}
%     f = p\circ r := \left( \langle p^1, r^1 \rangle,\dots, \langle p^1, r^C \rangle, \dots,  \langle p^K, r^1 \rangle,\dots, \langle p^K, r^C \rangle \right) \in \mathbb{R}^{CK},
% \end{equation}
% where $\langle p^i, r^j \rangle$ denotes the Frobenius inner product of $p^i$ and $r^j$.
% The resulting representations $f$ are referred to as the geometric layout modulated representations, and will be fed to~\Cref{eq:task} for retrieving the best-matching aerial images.

% \subsubsection{Polar Transformation}
% {\color{red}TBD, maybe citing}

\subsubsection{Geometric Layout Extractor}
 
This sub-module mines the global contextual information among the visual features and produce effective geometric layout descriptors. Despite the change in appearance across views, the arrangement of visual features remains largely intact. Hence, integrating the geometric layout information into the latent representations $f^v$ would improve its discriminative power for cross-view geo-localization.
% The correspondence between the ground and aerial views undergoes restrict geometric constraints,
% % which determine the relation between the geometric layouts of the two views, $F:\mathcal{G}^g$ and $\mathcal{G}^a$,
% accounting for the deformation of visual features and how their arrangement in the image changes from one view to another. The position of visual features is the key for matching the query ground image to reference aerial images. By leveraging the polar transformation, the position of visual features can be coarsely aligned. However, due to lack of precise information to reconstruct the 3-D scene in one satellite image, visual features in polar transformed images are distorted. Then, our model aims to extract the positions of visual features.
Note that the geometric layout is a global property in the sense that it captures the spatial configuration of multiple visual features at different positions in the ground and aerial images. For example, a single visual feature can span across the image, such as the road.
As a result, the geometric layout descriptors should be able to grasp the global correlation among visual features.

%To achieve this global layout understanding, 
In order to accomplish this goal, the geometric layout extractor is built on top of the standard transformer.
As shown in~\Cref{fig:model} \xh{(b)}, the network consists of a max-pooling layer along channels, a transformer module, and two embedding layers that locate before and after the transformer. The max-pooling layer produces a saliency map $M \in \mathbb{R}^{H \times W}$. Then, $M$ is processed by the first embedding layer, projected into $K$ embedding vectors $E=[e_1,e_2,...,e_{k}]$ \xh{where for each embedding vector the dimension is $(H \times W)/2$}. On top of the standard learnable positional embedding (PE) \xh{$E_{pe}$}~\citep{ViT}, we introduce an extra index-aware positional embedding adding to $E$
\begin{equation}
    \hat{E} = E+\text{HardTanh}(\xh{E_{pe}}+W_{LN}M^{idx}),
\end{equation}
where $M^{idx}_{{\xh{m}}j} = \frac{1}{C}{\arg\max}_{c\in 1,\dots,C}\, r^c_{{\xh{m}}j}$. \xh{$W_{LN}$ is a learnable linear transform that maps $M^{idx}$ to $K$ distinct subspaces, $W_{LN}\in \mathbb{R}^{(H\times W) \times (K \times \frac{H \times W}{2})}$.} The transformer explores correlations among $\hat{E}$. After projecting by the second embedding layer,
our geometric layout extractor generates a set of $K$ geometric layout descriptors $\mathbf{p}$.
The detailed model settings can be found in the supplementary material.

\subsubsection{Counterfactual-based Learning Schema}
% \begin{wrapfigure}[12]{l}{0.4\textwidth}
%   \begin{center}
%     \includegraphics[page=6,width=0.4\textwidth, clip,trim=14cm 35cm 76cm 73cm]{figures/nips_figures.pdf}
%   \end{center}
%   \caption{Causal graph of the proposed counterfactual learning progress.}
%   \label{fig:causal}
% \end{wrapfigure}
% Without any constraint on learned descriptors, the model might learn trivial solutions. However, there is no ground truth geometric descriptors available. To tackle this problem, 
Due to the absence of ground truth geometric layout descriptors, the sub-module $\mathcal{G}^v(\cdot)$ would only receive indirect and insufficient supervision during training. 
Inspired by~\citep{fine_grained_cf}, we propose 
% to tackle this problem by employing 
a counterfactual-based (CF-based) learning process.
% adopt the counterfactual intervention in the deep metric learning framework. 
% The causal graph is shown in~\Cref{fig:causal}. 
Specifically, we apply an intervention $do(\mathbf{p}^v=\hat{\mathbf{p}}^v)$ which substitutes $\mathbf{p}^v$ for a set of imaginary layout descriptors $\hat{\mathbf{p}}^v$ in~\Cref{eq:main}. This results in an imaginary representation $\hat{f}^v$.
Elements of $\hat{\mathbf{p}}^v$ are drawn from the uniform distribution $U[-1,1]$. 
% Then, an imaginary representation $\hat{f}$ is produced through substituting $\hat{\mathbf{p}}^{g(a)}$ for $\mathbf{p}^{g(a)}$ in~\Cref{eq:main}. Heuristically, the distance between $f$ and $\hat{f}$ reflects the quality of the realistic geometric descriptors $\mathbf{p}$. 
% Since, $\hat{p}$ is a random tensor without any geometric information from the input image. 
In order to penalize $\hat{\mathbf{p}}^v$ and encourage $\mathbf{p}^v$ to capture more distinctive geometric clues, we maximize the distance between $f^v$ and $\hat{f}^v$ \xh{by minimizing our proposed counterfactual loss}
\begin{equation}
\label{eq:cf}
    L^v_{cf} = \log\left(1+e^{-\beta^v \left[ d\left(f^v,\hat{f}^v\right) \right]}\right),
\end{equation}
where $\beta^v$ is a parameter to tune the convergence rate. \xh{The counterfactual loss provides a weakly supervision signal to the layout descriptors $p$ via penalizing the imaginary descriptors $\hat{p}$. In this way, the model can be away from apparently “wrong” solution and learn a better latent feature representation. } Besides the counterfactual loss, we also adopt the weighted soft margin triplet loss
% Counterfactual loss is not enough to train the model to learn the correspondence. We also adopt the weighted soft margin triplet loss which pushes the matched pairs closer and unmatched pairs further,
which pushes the matched pairs closer and unmatched pairs further away from each other
\begin{equation}
\label{eq:smtl}
    L_{triplet} = \log{\left( 1 + e^{\alpha\left[d(f^g_m, f^a_m)-d(f^g_m, f^a_n)\right]} \right)},
\end{equation}
where $\alpha$ is a hyperparameter that controls the convergence of training. $m,n \in \{1,2, \dots, N\}$ and $m \neq n$. Our final loss is
\begin{equation}
    L=L_{triplet}+L^a_{cf}+L^g_{cf}.
\end{equation}

\subsection{Layout Simulation and Semantic Augmentation}

In this paper, we elaborately design two categories of augmentations, i.e., Layout simulation and Semantic augmentation (\aug{}) to help improve the quality of extracted layout descriptors and the generalization of cross-view geo-localization models.

\begin{figure}[!t]
  \begin{center}
    \includegraphics[page=6,width=0.47\textwidth, clip,trim=6.5cm 70cm 63cm 2cm]{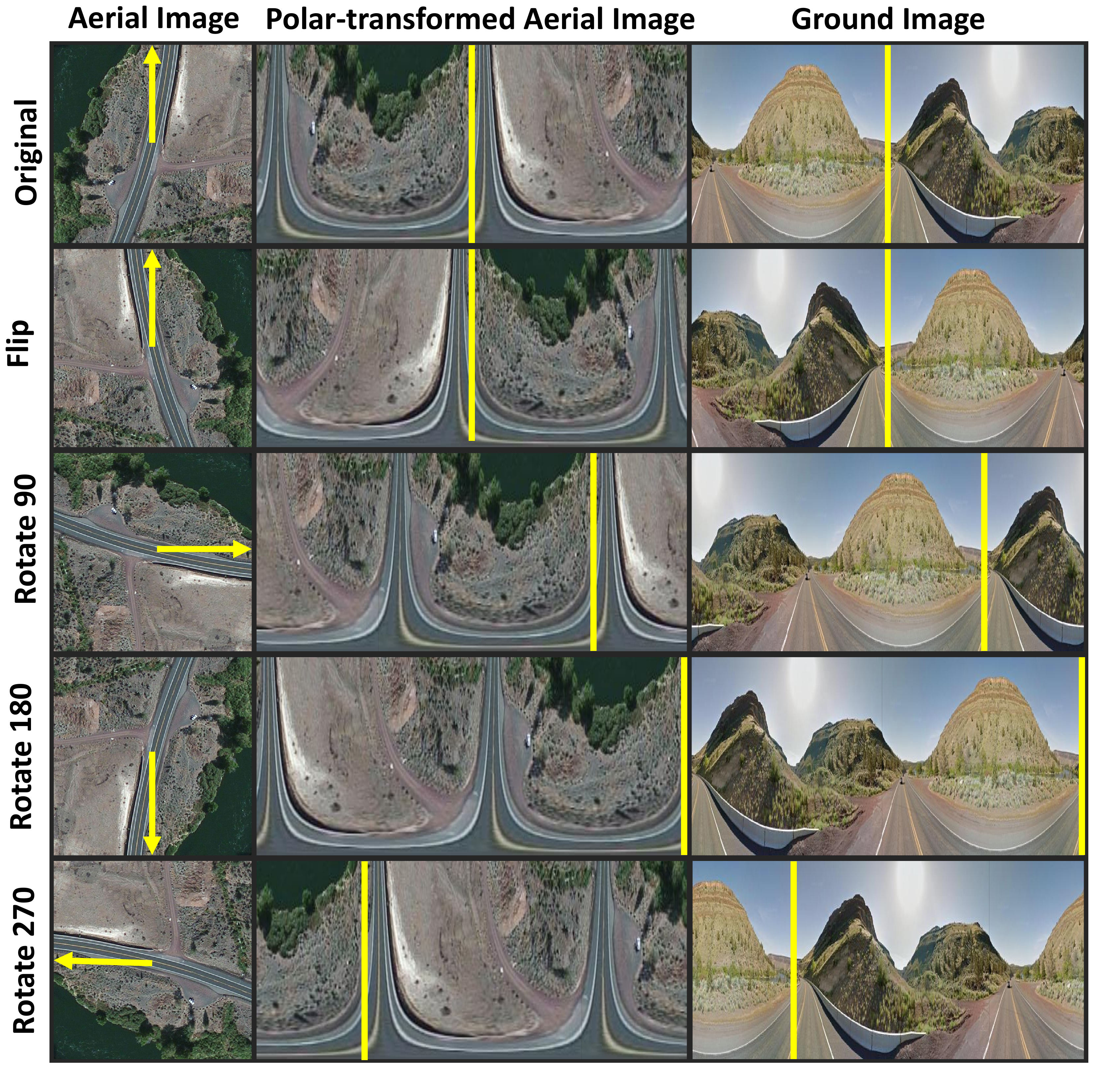}
  \end{center}
  \caption{Illustration of the layout simulation. From left to right are aerial image, polar transformed aerial image, and ground image. The yellow arrows and lines indicate the north direction.}
  \label{fig:layout_sim}
\end{figure}

% \noindent\textbf{Layout simulation:} 
\subsubsection{Layout Simulation}
It is a combination of a random flip and a random rotation ($90^{\circ}, 180^{\circ}, \text{ or } 270^{\circ}$)
% set of random combinations of flip, rotation $90^{\circ}, 180^{\circ}, \text{ and } 270^{\circ}$ 
that \textit{synchronously} applies to ground truth aerial and ground images. In this manner, low-level details are maintained, but the geometric layout is modified. As illustrated in \Cref{fig:layout_sim}, layout simulation can produce matched aerial-ground pairs with a different geometric layout.
% that does not exist in the real world.

% \noindent\textbf{Semantic augmentation:} 
\subsubsection{Semantic Augmentation}
Semantic augmentation randomly modifies the low-level features in aerial and ground images \textit{separately}. We employ color jitter to modify the brightness, contrast, and saturation in images. Moreover, we also randomly apply Gaussian blur and randomly transform images to grayscale images or posterized images. 
% converting-to-grayscale

Note that unlike previous data augmentation methods, our \aug{} does not break the geometric correspondence among visual features in the two views. Our experiments show that \aug{} greatly improves \ourmodel{} on cross-area performance while hardly weakening the same-area performance. Applying \aug{} to the existing methods also improves their performance in cross-area experiments. For more details on this, refer to the supplementary material.

% \input{tables/same_area_act}

% \input{tables/same_area_usa}
% Data augmentation is seldom used in cross-view geo-localization. 
% Traditional data augmentation methods, such as random crop, random erasing, typically undermine the alignment between images from the two views. 
% On the other hand, such an alignment turns out to be beneficial for improving model performance in cross-view geo-localization.
% In this paper, we elaborately design two categories of augmentations that 

\section{Experimental results}
\label{sec::experiment}
% \begin{table*}[!t]
%     \label{tab:same_area_usa}
%     \centering
%     \small
%     \setlength{\tabcolsep}{0.7 mm}
%     % \begin{tabular}{c|c||c|c|c|c}
%     \begin{tabular}{cccccccccc}
%     \toprule 
%     % \midrule
%     \multicolumn{5}{c}{w/o Polar Transformation (PT)} &
%     \multicolumn{5}{c}{w/ Polar Transformation (PT)} \\
%     \cmidrule(lr){1-5}
%     \cmidrule(lr){6-10}
%     Method & \Rone{} & \Rfive{} & \Rten{} & \Ronep & Method & \Rone{} & \Rfive{} & \Rten{} & \Ronep \\
%     \midrule
%     Regmi~\citep{bridging} & 48.75\% & - & 81.27\% & 95.98\%   &  SAFA~\citep{SAFA} & 89.84\% & 96.93\% & 98.14\% & 99.64\% \\
%     CVFT~\citep{featureTransport} & 61.43\% & 84.69\% & 90.49\% & 99.02\%  &   DSM~\citep{DSM} & 91.93\% & 97.50\% & 98.54\% & 99.67\% \\
%     SAFA~\citep{SAFA} & 81.15\% & 94.23\% & 96.85\% & 99.49\%   & Tocker~\citep{comingDown} & 92.56\% & 97.55\% & 98.33\% & 99.57\% \\
%     L2LTR~\citep{l2ltr} & \ST{91.99\%} & \ST{97.68\%} & \ST{98.65\%} & \ST{99.75\%}    &   L2LTR~\citep{l2ltr} & \ST{94.05\%} & \ST{98.27\%} & \ST{98.99\%} & \ST{99.67\%} \\
%     TransGeo~\citep{transgeo} & 94.08\% & 98.36\% & 99.04\% & 99.77\%    &   \ourmodel{}(Ours) w/ \aug{} & \FT{95.43\%} &	\FT{98.86\%} &	\FT{99.34\%} &	\FT{99.86\%} \\
%     \ourmodel{}(Ours) w/ \aug{} & \FT{93.76\%} &	\FT{98.47\%} &	\FT{99.22\%} &	\FT{99.85\%} \\
%     \bottomrule
%     \end{tabular}
%     \caption{Comparison between our \ourmodel{} w/ \aug{} and existing methods on CVUSA~\citep{CVUSA} dataset. 
% % ``PT'' stands for polar transformation.
% \FT{Best} results are shown in bold.}
% \end{table*}

\begin{table}[!ht]
    \centering
    \begin{tabular}{ccccc}
    \toprule \toprule
         Method & \Rone{} & \Rfive{} & \Rten{} & \Ronep  \\
         \midrule
        %  CVM-Net & 22.47\% & 49.98\% & 63.18\% & 93.62\% \\
        %  OriCNN & 40.79\% & 66.82\% & 76.36\% & 96.12\% \\
         FusionGAN & 48.75\% & - & 81.27\% & 95.98\% \\
         CVFT & 61.43\% & 84.69\% & 90.49\% & 99.02\% \\
         SAFA & 81.15\% & 94.23\% & 96.85\% & 99.49\% \\
         SAFA$\dagger$ & 89.84\% & 96.93\% & 98.14\% & 99.64\% \\
         DSM$\dagger$ & 91.93\% & 97.50\% & 98.54\% & 99.67\% \\
         CDE$\dagger$ & 92.56\% & 97.55\% & 98.33\% & 99.57\% \\
         L2LTR & 91.99\% & 97.68\% & 98.65\% & 99.75\% \\
         L2LTR$\dagger$ & 94.05\% & 98.27\% & 98.99\% & 99.67\%\\
         TransGeo & 94.08\% & 98.36\% & 99.04\% & 99.77\% \\
         SEH$\dagger$ & \SB{95.11\%} & 98.45\% & 99.00\% & 99.78\% \\
         Ours w/ \aug{} & 93.76\% &	\SB{98.47\%} &	\SB{99.22\%} &	\SB{99.85\%} \\
         Ours w/ \aug{}$\dagger$ & \B{95.43\%} &	\B{98.86\%} &\B{99.34\%} &	\B{99.86\%} \\ \bottomrule \bottomrule
    \end{tabular}
    \caption{\label{tab:same_area_usa}Comparison between the proposed \ourmodel{} and baseline methods
    % FusionGAN~\citep{bridging}, CVFT~\citep{featureTransport}, SAFA~\citep{SAFA}, DSM~\citep{DSM}, Coming Down to Earth(CDE)~\citep{comingDown}, L2LTR~\citep{l2ltr}, TransGeo~\citep{transgeo}
    on CVUSA dataset. $\dagger$ represents that polar transformation is applied to aerial images. \B{Best} results shown in  magenta and \SB{second best} results shown in cyan.}
\end{table}

\begin{table*}[!ht]
    \centering
    % \setlength{\tabcolsep}{1.0 mm}
    % \begin{tabular}{c|c||c|c|c|c||c|c|c|c}
    \begin{tabular}{ccccccccc}
    \toprule 
    % \toprule
    \multirow{2}{*}{Method} & \multicolumn{4}{c}{CVACT\_val} &
    % \multicolumn{2}{c||}{} & \multicolumn{4}{c||}{CVACT\_val} &
    \multicolumn{4}{c}{CVACT\_test} \\
    \cmidrule(lr){2-5}
    \cmidrule(lr){6-9}
    % \midrule
    % \hhline{--||----||----}
    % \hline
     & \Rone{} & \Rfive{} & \Rten{} & \Ronep & \Rone{} & \Rfive{} & \Rten{} & \Ronep \\
    \midrule
    % \hhline{--||----||----}
    % \multirow{7}{*}{w/o} & CVM-Net & 20.15\% & 45.00\% & 56.87\% & 87.57\% & 5.41\% & 14.79\% & 25.63\% & 54.53\% \\
    % & OriCNN & 46.96\% & 68.28\% & 75.48\% & 92.01\% & 19.21\% & 35.97\% & 43.30\% & 60.69\% \\
    CVFT & 61.05\% & 81.33\% & 86.52\% & 95.93\% & 26.12\% & 45.33\% & 53.80\% & 71.69\% \\
    SAFA & 78.28\% & 91.60\% & 93.79\% & 98.15\% & - & - & - & - \\
    SAFA$\dagger$ & 81.03\% & 92.80\% & 94.84\% & 98.17\% & 55.50\% & 79.94\% & 85.08\% & 94.49\% \\
    DSM$\dagger$ & 82.49\% & 92.44\% & 93.99\% & 97.32\% & 35.63\% & 60.07\% & 69.10\% & 84.75\% \\
    CDE$\dagger$ & 83.28\% & 93.57\% & 95.42\% & 98.22\% & 61.29\% & 85.13\% & 89.14\% & 98.32\% \\
    L2LTR & 83.14\% & 93.84\% & 95.51\% & \SB{98.40\%} & 58.33\% & 84.23\% & 88.60\% & 95.83\% \\
    L2LTR$\dagger$ & 84.89\% & 94.59\% & 95.96\% & 98.37\% & 60.72\% & 85.85\% & 89.88\% & 96.12\% \\
    TransGeo & 84.95\% & 94.14\% & 95.78\% & 98.37\% & - & - & - & - \\
    SEH$\dagger$ & 84.75\% & 93.97\% & 95.46\% & 98.11\% & - & - & - & - \\
    Ours w/ LS & \SB{85.43\%} &	\SB{94.81\%} & \SB{96.11\%} & 98.26\% & \SB{62.96\%} & \SB{87.35\%} & \SB{90.70\%} & \SB{98.61\%} \\
    Ours w/ LS$\dagger$ & \B{86.21\%} &	\B{95.44\%} & \B{96.72\%} & \B{98.77\%} & \B{64.52\%} &	\B{88.59\%} & \B{91.96\%} &	\B{98.74\%} \\
    % \midrule
    \bottomrule
    \end{tabular}
    \caption{\label{tab:same_area_act}Comparison between our \ourmodel{} w/ \aug{} and baseline methods
    % CVFT~\citep{featureTransport}, SAFA~\citep{SAFA}, DSM~\citep{DSM}, Coming Down to Earth(CDE)~\citep{comingDown}, L2LTR~\citep{l2ltr}, TransGeo~\citep{transgeo}
    on CVACT dataset. Notations are the same as \cref{tab:same_area_usa}.
    % \B{Best} results shown in  magenta and \SB{second best} results shown in cyan. $\dagger$ represents the polar transformation is applied to aerial images. 
    }
\end{table*}

\subsection{Experiment Settings\label{sec:expt_setting}}
% \noindent\textbf{Dataset:}
\paragraph{Dataset.}
To evaluate the effectiveness of \ourmodel{}, we conduct extensive experiments on two datasets, CVUSA~\citep{CVUSA}, and CVACT~\citep{liu2019lending}.
%CVUSA~\citep{CVUSA} contains $35,532$ training pairs and $8,884$ testing pairs. Similar to CVUSA~\citep{CVUSA}, 
Both CVUSA and CVACT contain $35,532$ training pairs. CVUSA provides $8,884$ pairs for testing and CVACT has the same number of pairs in its validation set (CVACT\_val). Besides, CVACT provides a challenging and large-scale testing set (CVUSA\_test) which contains $92,802$ pairs.
In CVUSA, we identify $762$ and $43$ repeated pairs in the original training set and testing set, respectively. We remove the repeated pairs in the training set but keep the testing set unchanged for fair comparisons.
Please refer to the supplementary material for more information of the duplicate pairs in CVUSA dataset.

% \noindent\textbf{Evaluation Metric:}
\paragraph{Evaluation Metric.}
Similar to existing methods~\citep{SAFA,comingDown,l2ltr,cvmnet,liu2019lending,DSM}, we choose to use recall accuracy at top $K$ (R@$K$) for evaluation purpose. R@$K$ measures the probability of the ground truth aerial image ranking within the first $K$ predictions given a query image. In the following experiments, we evaluate the performance of all methods on \Rone{}, \Rfive{}, \Rten{}, and \Ronep{}.
% For more information of our implementation details \xh{(i.e. different latent feature dimensions)}, please refer to the supplementary material.
\xh{\paragraph{Implementation Detail.} We employ a ResNet-34~\citep{resnet} as the backbone for a fair comparison with other baselines. $\alpha$ and $\beta$ are set to $10$ and $5$ respectively. We train the model on a single Nvidia V100 GPU for $200$ epochs with AdamW~\citep{adamw} optimizer. For more information (i.e. LS techniques and latent feature dimensions, etc.), please refer to the supplementary material.}

% \noindent\textbf{Implementation:}
% We implement our proposed model in Pytorch~\citep{pytorch}. The ground images and aerial images are resized to $122 \times 671$ and $256 \times 256$, respectively. Similar to previous methods~\citep{SAFA,DSM,l2ltr,comingDown,cvmnet}, we set the batch size to $32$ and $\alpha$ to $10$ in \Cref{eq:smtl}. Within each batch, the exhaustive mini-batch strategy~\citep{Vo} is utilized for constructing triplet pairs. We use AdamW~\citep{adamw} to train our algorithm for $200$ epochs with weight decay of $0.03$ on a single Nvidia V100 GPU. The learning rate is chosen to be $10^{-4}$ with the cosine learning rate schedule. The $\beta$ in \Cref{eq:cf} is set to $5$. The backbone ResNet34~\citep{resnet} is pretrained on ImageNet~\citep{imagenet}. We adopt a $2$-layer transformer with $4$ heads for the geometric layout extractor which is initialized randomly. The number of geometric descriptors $K$ is set to $8$.

%All the experiments in the following sections are configured as above unless otherwise specified.

\subsection{Same-area Experiment}

We first evaluate \ourmodel{} on the same-area cross-view geo-localization tasks in which training and testing data are captured from the same region. The results on CVUSA~\citep{CVUSA} and CVACT~\citep{liu2019lending} benchmarks are shown in~\Cref{tab:same_area_usa} and~\Cref{tab:same_area_act}, respectively. For a fair and complete comparison, we present the performance of \ourmodel{} trained with and without polar transformation (PT) on aerial images. Specifically, in the CVUSA experiments (\Cref{tab:same_area_usa}), \xh{with PT, \ourmodel{} achieves the state-of-the-art (SOTA) result}. Without PT, our GeoDTR exceeds TransGeo~\citep{transgeo} on \Rfive{}, \Rten{}, and \Ronep{} and achieve comparable results on \Rone{}.

% Specifically, in the CVUSA experiments (\Cref{tab:same_area_usa}), \ourmodel{} exceeds all existing methods. More importantly, we improve the best \Rone{} performance from $91.99\%$ to $93.76\%$ without polar transformation and from $94.05\%$ to $95.43\%$ with polar transformation. 
As shown in~\Cref{tab:same_area_act}, \ourmodel{} also achieves substantial improvement in performance on CVACT. To be noticed, \ourmodel{} achieves $64.52\%$ on \Rone{} of CVACT\_test which is a \textit{3.23\%} increase from previous \xh{ SOTA (CDE~\citep{comingDown})} on this highly challenging benchmark. Furthermore, we also observe that when training without polar transformation, \ourmodel{} only suffers a minor decrease in performance ($1.67\%$ on \Rone{} of CVUSA, $0.78\%$ on \Rone{} of CVACT\_val, and $1.56\%$ on \Rone{} of CVACT\_test). We attribute this to the geometric layout descriptors that can adapt to the non-polar-transformed aerial inputs and capture the spatial configuration. More qualitative analyses on geometric layout descriptors are discussed in the later sections. The results of same-area experiments demonstrate the superiority of \ourmodel{}.
%In addition, our model without Layout simulation and Semantic augmentation (LS) achieve the state-of-the-art result on CVUSA benchmark. Noticeably, when training with LS, \ourmodel{} can even achieve better results. The comparison on CVACT benchmark is presented in~\Cref{tab:same_area_act}. On CVACT\_val, \ourmodel{} also achieve state-of-the-art results. On more challenging CVACT\_test, \ourmodel{} improve the \Rone{} from $58.33\%$~\citep{l2ltr} to $64.52\%$ (with LS) and $68.20\%$ (without LS). 
%More experimental comparisons are reported in the supplementary material that compare our proposed method with the existing methods using the our proposed augmentation techniques. 
% To summarize, Our proposed model showed the state-of-the-art on most of the same-area benchmarks with/without polar transformation.
%same-area experiments demonstrate the effectiveness of \ourmodel{}. It also shows that training with LS would benefit the model to mine more correspondent visual features.
% \input{tables/same_area_act}

\subsection{Cross-area Experiment}

\begin{table}
% \begin{table}
    \centering
    \small
    \setlength{\tabcolsep}{0.7 mm}
    % \begin{tabular}{c|c|c|c|c|c}
    \begin{tabular}{cccccc}
    \toprule
    % \toprule
    Model & Task  & \Rone{} & \Rfive{} & \Rten{} & \Ronep{} \\
    \midrule
    SAFA$\dagger$ & \multirow{6}{*}{
        \begin{tabular}{@{}c@{}}CVUSA\\ $\boldsymbol{\downarrow}$ \\ CVACT\end{tabular}
    } & 30.40\% & 52.93\% & 62.29\% & 85.82\% \\
    DSM$\dagger$ & &33.66\% & 52.17\% & 59.74\% & 79.67\%\\
    L2LTR$\dagger$ & & \SB{47.55\%} & \SB{70.58\%} & \SB{77.39\%} & 91.39\% \\
    TransGeo & & 37.81\% & 61.57\% & 69.86\% & 89.14\% \\
    Ours w/ LS  & & 43.72\% & 66.99\% & 74.61\% & \SB{91.83\%} \\
    Ours w/ LS$\dagger$  & & \B{53.16\%} & \B{75.62\%} & \B{81.90\%} & \B{93.80\%} \\ \midrule
    SAFA$\dagger$ & \multirow{6}{*}{
        \begin{tabular}{@{}c@{}}CVACT\\ $\boldsymbol{\downarrow}$ \\ CVUSA\end{tabular}
    } & 21.45\% & 36.55\% & 43.79\% & 69.83\% \\
    DSM$\dagger$ & & 18.47\% & 34.46\% & 42.28\% & 69.01\% \\
    L2LTR$\dagger$ & & \SB{33.00\%} & \SB{51.87\%} & \SB{60.63\%} & \SB{84.79\%} \\
    TransGeo & & 17.45\% & 32.49\% & 40.48\% & 69.14\% \\
    Ours w/ LS & & 29.85\% & 49.25\% & 57.11\% & 82.47\% \\
    Ours w/ LS$\dagger$ & & \B{44.07\%} & \B{64.66\%} & \B{72.08\%} & \B{90.09\%} \\ \bottomrule
    \end{tabular}
    \caption{\label{tab:cross_area}Comparison between \ourmodel{} w/ LS and baselines
    % SAFA~\citep{SAFA}, DSM~\citep{DSM}, L2LTR~\citep{l2ltr}, TransGeo~\citep{transgeo}
    on cross-area benchmarks. Notations are the same as \cref{tab:same_area_usa}.
    % $\dagger$ represents that polar transformation is applied during training. \B{Best} results shown in magenta and \SB{second best} results shown in cyan.
    }
    
\end{table}

To further evaluate the generalization of \ourmodel{} on unseen scenes, we conduct the cross-area experiments, i.e., training on CVUSA while testing on CVACT (CVUSA $\rightarrow$ CVACT) and vice versa (CVACT $\rightarrow$ CVUSA). The results are summarized in~\Cref{tab:cross_area}. On CVUSA $\rightarrow$ CVACT, \ourmodel{} achieves $53.16\%$ on \Rone{} which significantly exceeds the current \xh{SOTA}~\citep{l2ltr}. Since CVACT is densely sampled from a single city, its images might share more common visual features. Hence, we consider CVACT $\rightarrow$ CVUSA to be a more challenging task. We observe that \ourmodel{} outperforms all other methods by a substantial amount. 
From the analyses in later sections, this significant improvement in cross-area performance comes from the cooperation between the geometric layout descriptors and our \aug{} technique.
The geometric layout descriptors efficiently grasp the spatial correlation among visual features and the \aug{} technique helps to alleviate the model's overfitting to low-level details.
% In which, \ourmodel{} can learn more diverse layouts and independent from low-level details. In later sections, we also observe the universal of LS which improve the cross-area performance of other models.

\subsection{Qualitative Study of Geometric Descriptors}

% \subsubsection{Qualitative study of geometric descriptors}

\begin{figure*}[!t]
    \centering
    \includegraphics[width=0.96\textwidth]{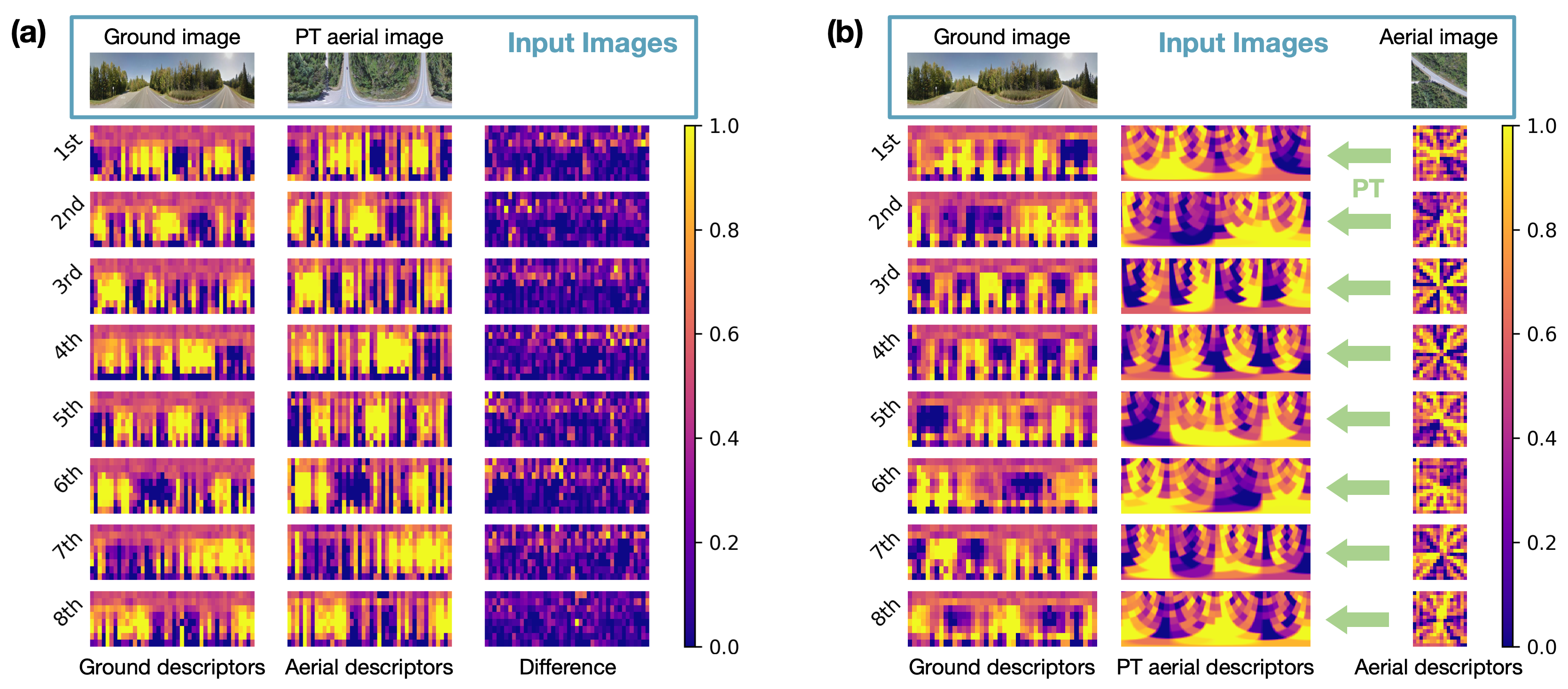}
    \caption{\label{fig:visualization} Visualization of learned descriptors from \ourmodel{} trained \textit{with} (a) and \textit{without} (b) polar transformation. Notice the \textit{alignment} between ground descriptors and aerial / PT aerial descriptors.}
\end{figure*}

% \begin{figure}[!t]
% \centering
% \begin{minipage}[t]{.49\textwidth}
%   \centering
%   \includegraphics[ scale=0.5]{figures/fig3.png}
% %   \captionof{figure}
%   \caption{Visualization of descriptors from \ourmodel{} trained \textit{with} polar transformation. 
%   %The input images are shown at the top. The descriptors are shown in heat maps. From left to right are ground descriptors, aerial descriptors, and the difference between them. From the difference where most region are close to $0$, it illustrates a strong alignment between the ground and aerial descriptors.
%   }
%   \label{fig:PT_vis}
% \end{minipage}%
% \hfill
% \begin{minipage}[t]{.49\textwidth}
%   \centering
%   \includegraphics[ scale=0.5]{figures/fig4.png}
% % %   \captionof{figure}
% %   \caption{Visualization of learned descriptors from \ourmodel{} trained \textit{without} polar transformation. For better visualization purpose, we then apply polar transform on aerial descriptors (right column) to generate polar transformed aerial descriptors (middle column). An alignment in patterns between ground descriptors (left column) and polar transformed aerial descriptors can be observed.}
%   \caption{Visualization of descriptors from \ourmodel{} trained \textit{without} polar transformation.}
%   \label{fig:noPT_vis}
% \end{minipage}
% \end{figure}

As discussed in our description of geometric layout extractor, our model learns to capture the geometric layout and then produces modulated latent representations upon that. The geometric layout could be considered as a fingerprint of the ground (aerial) image and should remain mostly the same under view-change. Consequently, for a ground-aerial pair, one would expect to see similar patterns in their geometric layout descriptors. To justify this point and fully demonstrate the power of our model, 
we visualize the ground and aerial descriptors in \Cref{fig:visualization} for cases when \ourmodel{} is trained with polar transformed aerial images and normal aerial images (without polar transformation), respectively.
% \Cref{fig:PT_vis} visualizes the ground and aerial descriptors predicted by \ourmodel{} training with polar transformed aerial images. Also,  \Cref{fig:noPT_vis} visualizes the descriptors predicted by \ourmodel{} training with normal aerial images (without polar transformation).

In~\Cref{fig:visualization}(a), we first note a strong alignment between descriptors of a given ground-aerial pair.
To better visualize, we also present the difference between the corresponding descriptors in the third column of the figure. It is clear to see that the corresponding descriptors possess very similar values apart from the ones located at a narrow strip near the top of each pair of descriptors.

% In~\Cref{fig:descriptos}(a), it is obvious that there is strong similarities between each pair of ground descriptors and aerial descriptors as shown by the difference column. 

More strikingly, such an alignment still exists when our model is trained with normal ground images. In~\Cref{fig:visualization}(b), we unroll the descriptors for the normal aerial image by polar transformation. We observe that apart from the deformation brought by the polar transformation, the locations of salient patterns in the aerial descriptors match those in the corresponding ground descriptors. This indicates the ability of \ourmodel{} to grasp the geometric correspondence even without the guidance of polar transformation.

\subsubsection{Ablation Study}

\begin{table}[!t]
\begin{subtable}[h]{0.45\textwidth}
\centering
\small
\setlength{\tabcolsep}{0.8 mm}
\begin{tabular}{ccccccc}
    \toprule 
    & \multicolumn{3}{c}{Same-area} &
    \multicolumn{3}{c}{Cross-area} \\
    \cmidrule(lr){2-4}
    \cmidrule(lr){5-7}
      & \Rone{} & \Rfive{} & \Ronep & \Rone{} & \Rfive{}  & \Ronep \\
    \midrule
    L2LTR &	94.05\% & 98.27\%  & 99.67\% & 47.55\% & 70.58\% & 91.39\% \\
       L2LTR$\ddag$ & 93.62\% & 98.46\%  & 99.77\% & 52.58\% & \B{75.81\%}  & 93.51\% \\[1mm]
       Ours & 95.23\% & 98.71\% & 99.79\% & 47.79\% & 70.52\%  & 92.20\% \\
       Ours$\ddag$ & \B{95.43\%} & \B{98.86\%}  &  \B{99.86\%} & \B{53.16\%} & 75.62\%  & \B{93.80\%} \\
    \bottomrule
\end{tabular}
\caption{Models are trained on CVUSA dataset.}
\end{subtable}

\begin{subtable}[h]{0.45\textwidth}
\centering
\small
\setlength{\tabcolsep}{0.8 mm}
\begin{tabular}{ccccccc}
    \toprule 
    & \multicolumn{3}{c}{Same-area} &
    \multicolumn{3}{c}{Cross-area} \\
    \cmidrule(lr){2-4}
    \cmidrule(lr){5-7}
      & \Rone{} & \Rfive{} & \Ronep & \Rone{} & \Rfive{}  & \Ronep \\
    \midrule
       L2LTR & 84.89\% & 94.59\%  & 98.37\% & 33.00\% & 51.87\%  & 84.79\% \\
       L2LTR$\ddag$ & 83.49\% & 94.93\% & 98.68\% & 37.69\% & 57.78\%  & 89.63\% \\[1mm]
       Ours & \B{87.42\%} & 95.37\%  & 98.65\% & 29.13\% & 47.86\% & 81.09\% \\
       Ours$\ddag$ & 86.21\% & \B{95.44\%}  & \B{98.77\%} & \B{44.07\%} & \B{64.66\%} & \B{90.09\%} \\
    \bottomrule
\end{tabular}
\caption{Models are trained on CVACT dataset.}
\end{subtable}
\caption{
Comparison between L2LTR~\citep{l2ltr} and \ourmodel{} with or without the proposed \aug{} technique. Results are shown for both cases when the models are trained on CVUSA and CVACT datasets. $\ddag$ represents that LS is applied during training. \B{Best} results are shown in magenta.
}
\label{tab:ablation_aug_othermethods}
\end{table}
\paragraph{\aug{} Technique}
It is worth emphasizing that our \aug{} stands as a generic technique to improve the generalization of cross-view geo-localization models. 
To illustrate this point, 
we apply the \aug{} technique to 
%SAFA~\citep{SAFA} which is a popular model in this research direction and
L2LTR~\citep{l2ltr} by comparing the recall accuracy training with or without \aug{}.
The comparison is presented in~\Cref{tab:ablation_aug_othermethods}, which also includes the same ablation on our \ourmodel{}. We notice that \aug{} greatly boosts the performance of L2LTR on the cross-area benchmark with only a minor decrease on the same-area benchmark. This indicates the effectiveness of \aug{} to improve cross-view geo-localization models in capturing clues for geo-localizing in unseen scenes.
%\aug{} could help improving the ability of the cross-view geo-localization models on unseen scenes.
% Also, we observe minor decrease of SAFA~\citep{SAFA} and L2LTR~\citep{l2ltr} on same area experiment.
% \Cref{fig:rel_diff} summarizes the results. 
% Both models achieve great boost in cross-area performance, with only minor sacrificing the same-area performance.
Note that even with the benefits from \aug{}, \ourmodel{} w/ \aug{} still \textit{outperforms} L2LTR w/ \aug{} on the overall performance in both same-area and cross-area benchmarks. The complete comparison with other existing models \xh{(i.e. SAFA)} can be found in the supplementary material.
% For better visualizing the effect of \aug{}, we plot the difference of recall rate in~\Cref{fig:rel_diff}. It is clear to see the performance improvement brought by \aug{} on all three models.

% \subsubsection{Ablation on number of descriptors}
\begin{table}[!t]
\centering
\small
\setlength{\tabcolsep}{1.4 mm}
\begin{tabular}{cccccc}
    \toprule 
    % \multirow{2}{*}{} & \multirow{2}{*}{$N_{des.}$} &
    % \multicolumn{1}{l}{\textbf{Ablation}} 
     & Configuration & \Rone{} & \Rfive{} & \Rten{} & \Ronep \\
    \midrule
    \multirow{6}{*}{
        \begin{tabular}{@{}c@{}}
            \rotatebox[origin=c]{90}{\makecell{Same-area}}
        \end{tabular}
    } & (1, CF, w/ \aug{}) & 93.44\% & 98.18\% & 99.05\% & 99.80\% \\
    %   & (2, CF, w/ \aug{}) & 94.09\% & 98.44\% & 99.13\% & 99.80\% & 43.07\% & 67.30\% & 74.52\% & 90.32\% \\
      & (4, CF, w/ \aug{}) & 94.80\% & 98.64\% & 99.30\% & 99.82\% \\
      & (8, CF, w/ \aug{}) & 95.43\% & 98.86\% & 99.34\% & 99.86\% \\
      & (8, noCF, w/ \aug{}) & 95.06\% & 98.72\% & 99.30\% & 99.85\% \\
      & (8, noCF, w/o \aug{}) & 94.83\% & 98.59\% & 99.28\% & 99.80\% \\
      & (8, CF, w/o \aug{}) & 95.23\% & 98.71\% & 99.26\% & 99.79\% \\
    \midrule
    \multirow{6}{*}{
        \begin{tabular}{@{}c@{}}
            \rotatebox[origin=c]{90}{\makecell{Cross-area}}
        \end{tabular}
    } & (1, CF, w/ \aug{}) & 39.93\% & 63.54\% & 71.59\% & 88.92\% \\
    & (4, CF, w/ \aug{}) & 46.36\% & 69.37\% & 76.46\% & 91.27\% \\
    & (8, CF, w/ \aug{}) & 53.16\% & 75.62\% & 81.90\% & 93.80\% \\
    & (8, noCF, w/ \aug{}) & 49.18\% & 71.96\% & 79.28\% & 92.84\% \\
    & (8, noCF, w/o \aug{}) & 43.71\% & 66.87\% & 74.68\% & 90.66\% \\
    & (8, CF, w/o \aug{}) & 47.79\% & 70.52\% & 77.52\% & 92.20\% \\ \bottomrule
    
\end{tabular}
\caption{
Performance of \ourmodel{} under different configurations, including
the number of descriptors $K$, with or without counterfactual loss (CF or noCF), and with or without \aug{} (w/ \aug{} or w/o \aug{}).
% on same-area and cross-area benchmarks. 
Results are shown for both cases when our model is trianed on CVUSA and CVACT datasets.}
\label{tab:ablation_descriptors}
\end{table}

\paragraph{Geometric Layout Descriptors}
To demonstrate the benefit of geometric layout descriptors to \ourmodel{}, we conduct ablation experiments with the different number of descriptors $K$. The results are included in the first three lines of the upper and bottom parts of the ablation study (\Cref{tab:ablation_descriptors}). 
% In the experiments, \ourmodel{} are trained with \aug{} augmentations.
%In the first three lines of the upper and bottom parts of the table,
We observe that with more descriptors, the performance is \textit{constantly} improved, especially the cross-area ones. This implies the importance of the geometric layout for the cross-area task.
%This implies that a high-quality modeling of geometric layout could be the key for the cross-area cross-view geo-localization task.
Moreover, there is a notable gap in the cross-area performance between $K=1$ and $K=8$ cases.
This gap highlights the substantial contribution of the geometric layout descriptors \textit{in addition} to the \aug{} technique, and, thus, reflects the effectiveness of our model design.
% the fact that adding more descriptors brings substantial improvements in cross-area performance 
% \subsubsection{Ablation on CF-based learning}
\paragraph{CF Learning Schema}
The effects of the CF-based learning process are shown in the last four lines in the upper and bottom parts of~\Cref{tab:ablation_descriptors}. 
We find that CF-based learning boosts the recall accuracy except for a few limited cases.
The improvement is more evident in the cross-area performance when our model is trained on the CVUSA dataset. To be noticed, the value of \Rone{} and \Rfive{} increase from $49.18\%$ to $53.16\%$ and from $71.96\%$ to $75.62\%$, respectively.
% \subsubsection{Computational Efficiency}
% In this section, we directly compare the number of parameters, inference time, and pretrained weight between \ourmodel{} and current state-of-the-art methods L2LTR~\citep{l2ltr}. The number of parameter of \ourmodel{} is only $48.5M$ which is $5\times$ less parameters than L2LT~\citep{l2ltr} ($195.9M$). The inference time for \ourmodel{} is $235ms$ and L2LTR~\citep{l2ltr} is $405ms$. We almost double the inference speed compared with L2LTR. Finally, \ourmodel{} only needs ResNet34~\citep{resnet} pretrained model for backbone and is randomly initialized for the other components. L2LTR~\citep{l2ltr} is initialized with weights from ViT~\citep{ViT} which is pretrained on ImageNet-21K.

% \subsection{A Summary of Ablation Study}
% ablation of \ourmodel{} trained with different levels of \aug{} augmentations;
% ablation of \ourmodel{} trained with and without polar transformation;
% ablation of \ourmodel{} trained with and without counterfactual loss;
% % others like ablation of different backbones, different transformer arch may be not worthy mentioning here.

% we compare the performance with and without applying the \aug{} technique to SAFA~\citep{SAFA}, L2LTR~\citep{l2ltr} as well as our \ourmodel{}.
% we apply the \aug{} technique to SAFA~\citep{SAFA} and L2LTR~\citep{l2ltr}

\section{Conclusion and future works\label{sec:conclusion}}
\label{sec::conclusion}
To address the challenges in cross-view geo-localization, we propose \ourmodel{} 
% In this paper, we have proposed a novel cross-view geo-localization model \ourmodel{}
which disentangles geometric layout from raw input features and better explores the spatial correlations among visual features. In addition, we introduce layout simulation and semantic augmentation which improve the generalization of \ourmodel{} and other existing cross-view geo-localization models. 
%Those can be applied to the existing cross-view geo-localization models and they show substantial enhancements on cross-area benchmarks.
Moreover, a novel counterfactual-based learning process is introduced to train \ourmodel{}. 
Extensive experiments demonstrate the superiority of \ourmodel{} on standard, fine-grained, and cross-area cross-view geo-localization tasks. 
%Furthermore, our proposed layout simulation and semantic augmentation can be helpful in improving existing models. 
% \ourmodel{} is limited in lack of explainability on geometric layout descriptors. We aim to research on explainability of  cross-view geo-localization models for future works.
Presently, the interpretation of geometric layout descriptors in our model has not been fully explored.
In the future, we will keep investigating their properties and work towards more explainable models.

\section{Acknowledgement}

This project has been supported in part by VTrans and NOAA Award No. NA22NWS4320003 (subaward no. A22-0303-S001). Computations were performed on the Vermont Advanced Computing, supported in part by NSF award No.OAC-1827314. 04. \\
X. Y. L. and Y. Z. were supported by grants from the National Science and Technology Innovation 2030 Project of China (2021ZD0202600) and the National Science Foundation of China (U22B2063).

\bibliography{aaai23}

\begin{thebibliography}{36}
\providecommand{\natexlab}[1]{#1}

\bibitem[{Abbasnejad et~al.(2020)Abbasnejad, Teney, Parvaneh, Shi, and
  Hengel}]{VQA_cf}
Abbasnejad, E.; Teney, D.; Parvaneh, A.; Shi, J.; and Hengel, A. v.~d. 2020.
\newblock Counterfactual Vision and Language Learning.
\newblock In \emph{Proceedings of the IEEE/CVF Conference on Computer Vision
  and Pattern Recognition (CVPR)}.

\bibitem[{Baradel et~al.(2020)Baradel, Neverova, Mille, Mori, and
  Wolf}]{cophy_cf}
Baradel, F.; Neverova, N.; Mille, J.; Mori, G.; and Wolf, C. 2020.
\newblock CoPhy: Counterfactual Learning of Physical Dynamics.
\newblock In \emph{International Conference on Learning Representations}.

\bibitem[{Byrne(2019)}]{explain_cf}
Byrne, R.~M. 2019.
\newblock Counterfactuals in Explainable Artificial Intelligence (XAI):
  Evidence from Human Reasoning.
\newblock In \emph{IJCAI}, 6276--6282.

\bibitem[{Cai et~al.(2019)Cai, Guo, Khan, Hu, and Wen}]{hardTriplet}
Cai, S.; Guo, Y.; Khan, S.; Hu, J.; and Wen, G. 2019.
\newblock Ground-to-Aerial Image Geo-Localization With a Hard Exemplar
  Reweighting Triplet Loss.
\newblock In \emph{Proceedings of the IEEE/CVF International Conference on
  Computer Vision (ICCV)}.

\bibitem[{Chiu et~al.(2018)Chiu, Murali, Villamil, Kessler, Samarasekera, and
  Kumar}]{AR1}
Chiu, H.-P.; Murali, V.; Villamil, R.; Kessler, G.~D.; Samarasekera, S.; and
  Kumar, R. 2018.
\newblock Augmented Reality Driving Using Semantic Geo-Registration.
\newblock In \emph{2018 IEEE Conference on Virtual Reality and 3D User
  Interfaces (VR)}, 423--430.

\bibitem[{Deng et~al.(2009)Deng, Dong, Socher, Li, Li, and Fei-Fei}]{imagenet}
Deng, J.; Dong, W.; Socher, R.; Li, L.-J.; Li, K.; and Fei-Fei, L. 2009.
\newblock ImageNet: A large-scale hierarchical image database.
\newblock In \emph{2009 IEEE Conference on Computer Vision and Pattern
  Recognition}, 248--255.

\bibitem[{Dosovitskiy et~al.(2021)Dosovitskiy, Beyer, Kolesnikov, Weissenborn,
  Zhai, Unterthiner, Dehghani, Minderer, Heigold, Gelly, Uszkoreit, and
  Houlsby}]{ViT}
Dosovitskiy, A.; Beyer, L.; Kolesnikov, A.; Weissenborn, D.; Zhai, X.;
  Unterthiner, T.; Dehghani, M.; Minderer, M.; Heigold, G.; Gelly, S.;
  Uszkoreit, J.; and Houlsby, N. 2021.
\newblock An Image is Worth 16x16 Words: Transformers for Image Recognition at
  Scale.
\newblock In \emph{International Conference on Learning Representations}.

\bibitem[{Goodfellow et~al.(2014)Goodfellow, Pouget-Abadie, Mirza, Xu,
  Warde-Farley, Ozair, Courville, and Bengio}]{gan}
Goodfellow, I.; Pouget-Abadie, J.; Mirza, M.; Xu, B.; Warde-Farley, D.; Ozair,
  S.; Courville, A.; and Bengio, Y. 2014.
\newblock Generative adversarial nets.
\newblock \emph{Advances in neural information processing systems}, 27.

\bibitem[{Guo et~al.(2022)Guo, Choi, Li, Boussaid, and Bennamoun}]{SEH}
Guo, Y.; Choi, M.; Li, K.; Boussaid, F.; and Bennamoun, M. 2022.
\newblock Soft Exemplar Highlighting for Cross-View Image-Based
  Geo-Localization.
\newblock \emph{IEEE transactions on image processing}, 31: 2094--2105.

\bibitem[{He et~al.(2016)He, Zhang, Ren, and Sun}]{resnet}
He, K.; Zhang, X.; Ren, S.; and Sun, J. 2016.
\newblock Deep Residual Learning for Image Recognition.
\newblock In \emph{Proceedings of the IEEE Conference on Computer Vision and
  Pattern Recognition (CVPR)}.

\bibitem[{Hu et~al.(2018)Hu, Feng, Nguyen, and Lee}]{cvmnet}
Hu, S.; Feng, M.; Nguyen, R.~M.; and Lee, G.~H. 2018.
\newblock Cvm-net: Cross-view matching network for image-based ground-to-aerial
  geo-localization.
\newblock In \emph{Proceedings of the IEEE Conference on Computer Vision and
  Pattern Recognition}, 7258--7267.

\bibitem[{Kim and Walter(2017)}]{kim2017satellite}
Kim, D.-K.; and Walter, M.~R. 2017.
\newblock Satellite image-based localization via learned embeddings.
\newblock In \emph{2017 IEEE International Conference on Robotics and
  Automation (ICRA)}, 2073--2080.

\bibitem[{Lin, Belongie, and Hays(2013)}]{lin2013cross}
Lin, T.-Y.; Belongie, S.; and Hays, J. 2013.
\newblock Cross-View Image Geolocalization.
\newblock In \emph{Proceedings of the IEEE Conference on Computer Vision and
  Pattern Recognition (CVPR)}.

\bibitem[{Lin et~al.(2015)Lin, Cui, Belongie, and Hays}]{lin2015learning}
Lin, T.-Y.; Cui, Y.; Belongie, S.; and Hays, J. 2015.
\newblock Learning Deep Representations for Ground-to-Aerial Geolocalization.
\newblock In \emph{Proceedings of the IEEE Conference on Computer Vision and
  Pattern Recognition (CVPR)}.

\bibitem[{Liu and Li(2019)}]{liu2019lending}
Liu, L.; and Li, H. 2019.
\newblock Lending Orientation to Neural Networks for Cross-View
  Geo-Localization.
\newblock In \emph{Proceedings of the IEEE/CVF Conference on Computer Vision
  and Pattern Recognition (CVPR)}.

\bibitem[{Loshchilov and Hutter(2017)}]{adamw}
Loshchilov, I.; and Hutter, F. 2017.
\newblock Decoupled weight decay regularization.
\newblock \emph{arXiv preprint arXiv:1711.05101}.

\bibitem[{Paszke et~al.(2019)Paszke, Gross, Massa, Lerer, Bradbury, Chanan,
  Killeen, Lin, Gimelshein, Antiga, Desmaison, Kopf, Yang, DeVito, Raison,
  Tejani, Chilamkurthy, Steiner, Fang, Bai, and Chintala}]{pytorch}
Paszke, A.; Gross, S.; Massa, F.; Lerer, A.; Bradbury, J.; Chanan, G.; Killeen,
  T.; Lin, Z.; Gimelshein, N.; Antiga, L.; Desmaison, A.; Kopf, A.; Yang, E.;
  DeVito, Z.; Raison, M.; Tejani, A.; Chilamkurthy, S.; Steiner, B.; Fang, L.;
  Bai, J.; and Chintala, S. 2019.
\newblock PyTorch: An Imperative Style, High-Performance Deep Learning Library.
\newblock In Wallach, H.; Larochelle, H.; Beygelzimer, A.; d\textquotesingle
  Alch\'{e}-Buc, F.; Fox, E.; and Garnett, R., eds., \emph{Advances in Neural
  Information Processing Systems 32}, 8024--8035. Curran Associates, Inc.

\bibitem[{Pearl(2009)}]{Pearl_Causality}
Pearl, J. 2009.
\newblock \emph{Causality: Models, Reasoning and Inference}.
\newblock USA: Cambridge University Press, 2nd edition.
\newblock ISBN 052189560X.

\bibitem[{Rao et~al.(2021)Rao, Chen, Lu, and Zhou}]{fine_grained_cf}
Rao, Y.; Chen, G.; Lu, J.; and Zhou, J. 2021.
\newblock Counterfactual Attention Learning for Fine-Grained Visual
  Categorization and Re-Identification.
\newblock In \emph{Proceedings of the IEEE/CVF International Conference on
  Computer Vision (ICCV)}, 1025--1034.

\bibitem[{Regmi and Shah(2019)}]{bridging}
Regmi, K.; and Shah, M. 2019.
\newblock Bridging the Domain Gap for Ground-to-Aerial Image Matching.
\newblock In \emph{Proceedings of the IEEE/CVF International Conference on
  Computer Vision (ICCV)}.

\bibitem[{Rodrigues and Tani(2021)}]{rodrigues2021these}
Rodrigues, R.; and Tani, M. 2021.
\newblock Are These From the Same Place? Seeing the Unseen in Cross-View Image
  Geo-Localization.
\newblock In \emph{Proceedings of the IEEE/CVF Winter Conference on
  Applications of Computer Vision (WACV)}, 3753--3761.

\bibitem[{Rodrigues and Tani(2022)}]{Rodrigues_2022_WACV}
Rodrigues, R.; and Tani, M. 2022.
\newblock Global Assists Local: Effective Aerial Representations for Field of
  View Constrained Image Geo-Localization.
\newblock In \emph{Proceedings of the IEEE/CVF Winter Conference on
  Applications of Computer Vision (WACV)}, 3871--3879.

\bibitem[{Shetty and Gao(2019)}]{UAV1}
Shetty, A.; and Gao, G.~X. 2019.
\newblock UAV Pose Estimation using Cross-view Geolocalization with Satellite
  Imagery.
\newblock In \emph{2019 International Conference on Robotics and Automation
  (ICRA)}, 1827--1833.

\bibitem[{Shi et~al.(2019)Shi, Liu, Yu, and Li}]{SAFA}
Shi, Y.; Liu, L.; Yu, X.; and Li, H. 2019.
\newblock Spatial-aware feature aggregation for image based cross-view
  geo-localization.
\newblock \emph{Advances in Neural Information Processing Systems}, 32:
  10090--10100.

\bibitem[{Shi et~al.(2020{\natexlab{a}})Shi, Yu, Campbell, and Li}]{DSM}
Shi, Y.; Yu, X.; Campbell, D.; and Li, H. 2020{\natexlab{a}}.
\newblock Where Am I Looking At? Joint Location and Orientation Estimation by
  Cross-View Matching.
\newblock In \emph{Proceedings of the IEEE/CVF Conference on Computer Vision
  and Pattern Recognition (CVPR)}.

\bibitem[{Shi et~al.(2020{\natexlab{b}})Shi, Yu, Liu, Zhang, and
  Li}]{featureTransport}
Shi, Y.; Yu, X.; Liu, L.; Zhang, T.; and Li, H. 2020{\natexlab{b}}.
\newblock Optimal Feature Transport for Cross-View Image Geo-Localization.
\newblock \emph{Proceedings of the AAAI Conference on Artificial Intelligence},
  34(07): 11990--11997.

\bibitem[{Toker et~al.(2021)Toker, Zhou, Maximov, and Leal-Taixe}]{comingDown}
Toker, A.; Zhou, Q.; Maximov, M.; and Leal-Taixe, L. 2021.
\newblock Coming Down to Earth: Satellite-to-Street View Synthesis for
  Geo-Localization.
\newblock In \emph{Proceedings of the IEEE/CVF Conference on Computer Vision
  and Pattern Recognition (CVPR)}, 6488--6497.

\bibitem[{Vaswani et~al.(2017)Vaswani, Shazeer, Parmar, Uszkoreit, Jones,
  Gomez, Kaiser, and Polosukhin}]{attention}
Vaswani, A.; Shazeer, N.; Parmar, N.; Uszkoreit, J.; Jones, L.; Gomez, A.~N.;
  Kaiser, L.~u.; and Polosukhin, I. 2017.
\newblock Attention is All you Need.
\newblock In Guyon, I.; Luxburg, U.~V.; Bengio, S.; Wallach, H.; Fergus, R.;
  Vishwanathan, S.; and Garnett, R., eds., \emph{Advances in Neural Information
  Processing Systems}, volume~30. Curran Associates, Inc.

\bibitem[{Vo and Hays(2016)}]{Vo}
Vo, N.~N.; and Hays, J. 2016.
\newblock Localizing and Orienting Street Views Using Overhead Imagery.
\newblock In Leibe, B.; Matas, J.; Sebe, N.; and Welling, M., eds.,
  \emph{Computer Vision -- ECCV 2016}, 494--509. Cham: Springer International
  Publishing.
\newblock ISBN 978-3-319-46448-0.

\bibitem[{Wang et~al.(2021)Wang, Zheng, Yan, Zhang, Sun, Zheng, and
  Yang}]{wang2021each}
Wang, T.; Zheng, Z.; Yan, C.; Zhang, J.; Sun, Y.; Zheng, B.; and Yang, Y. 2021.
\newblock Each Part Matters: Local Patterns Facilitate Cross-view
  Geo-localization.
\newblock \emph{IEEE Transactions on Circuits and Systems for Video
  Technology}, 1--1.

\bibitem[{Wang et~al.(2019)Wang, Wan, Zhang, Bai, Cui, and Yu}]{reinforce_cf}
Wang, Y.; Wan, Y.; Zhang, C.; Bai, L.; Cui, L.; and Yu, P. 2019.
\newblock Competitive multi-agent deep reinforcement learning with
  counterfactual thinking.
\newblock In \emph{2019 IEEE International Conference on Data Mining (ICDM)},
  1366--1371. IEEE.

\bibitem[{Wilson et~al.(2021)Wilson, Zhang, Sultani, and Wshah}]{survey}
Wilson, D.; Zhang, X.; Sultani, W.; and Wshah, S. 2021.
\newblock Visual and Object Geo-localization: A Comprehensive Survey.
\newblock \emph{arXiv preprint arXiv:2112.15202}.

\bibitem[{Workman, Souvenir, and Jacobs(2015)}]{CVUSA}
Workman, S.; Souvenir, R.; and Jacobs, N. 2015.
\newblock Wide-Area Image Geolocalization With Aerial Reference Imagery.
\newblock In \emph{Proceedings of the IEEE International Conference on Computer
  Vision (ICCV)}.

\bibitem[{Yang, Lu, and Zhu(2021)}]{l2ltr}
Yang, H.; Lu, X.; and Zhu, Y. 2021.
\newblock Cross-view Geo-localization with Layer-to-Layer Transformer.
\newblock In Ranzato, M.; Beygelzimer, A.; Dauphin, Y.; Liang, P.; and Vaughan,
  J.~W., eds., \emph{Advances in Neural Information Processing Systems},
  volume~34, 29009--29020. Curran Associates, Inc.

\bibitem[{Zheng, Wei, and Yang(2020)}]{zheng2020university}
Zheng, Z.; Wei, Y.; and Yang, Y. 2020.
\newblock University-1652: A Multi-View Multi-Source Benchmark for Drone-Based
  Geo-Localization.
\newblock In \emph{Proceedings of the 28th ACM International Conference on
  Multimedia}, MM '20, 1395–1403. New York, NY, USA: Association for
  Computing Machinery.
\newblock ISBN 9781450379885.

\bibitem[{Zhu, Shah, and Chen(2022)}]{transgeo}
Zhu, S.; Shah, M.; and Chen, C. 2022.
\newblock TransGeo: Transformer Is All You Need for Cross-View Image
  Geo-Localization.
\newblock In \emph{Proceedings of the IEEE/CVF Conference on Computer Vision
  and Pattern Recognition (CVPR)}, 1162--1171.

\end{thebibliography}

\newpage
\section{Supplementary material}

In this supplementary material, we provide additional information about the datasets, implementation, and societal impact. We also include the complete results of the ablation study and more illustrations of the geometric layout descriptors. Our code can be found in the attached code folder with supplementary material.

\section{Dataset}
\subsection{Consent}
We obtain permission for the CVUSA dataset from the owner by submitting the MVRL Dataset Request Form~\footnote{\url{https://mvrl.cse.wustl.edu/datasets/cvusa/}}. We obtain the permission of the CVACT dataset by contacting the author directly.

\subsection{Repeated data in CVUSA}
\begin{figure*}[!ht]
    \centering
    \includegraphics[width=\textwidth]{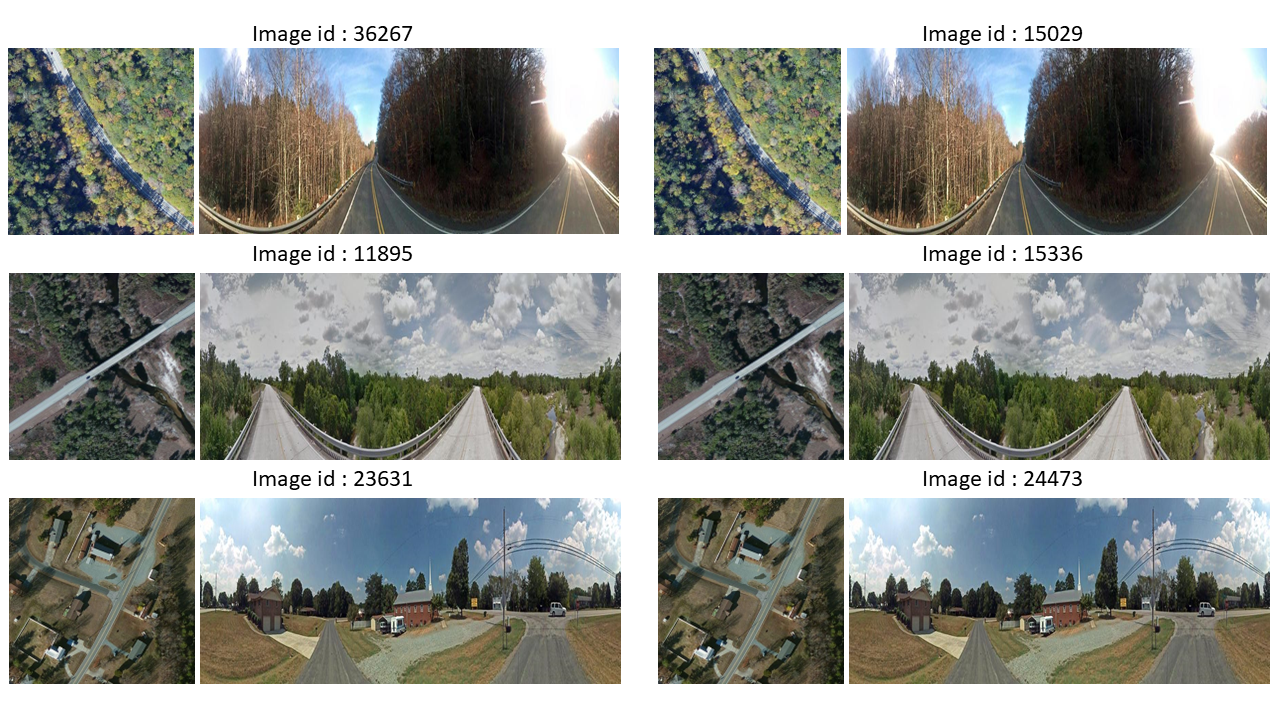}
    \caption{Three examples of repeated ground-aerial pairs from CVUSA training set.}
    \label{fig:duplicate}
\end{figure*}
As discussed in the main text, we identify $762$ and $43$ repeated pairs in the original training set and testing set of the CVUSA dataset, respectively. We apply the md5 algorithm on the pixel values of each image to identify these repeated pairs. We show three examples in \Cref{fig:duplicate}.

\section{Societal Impact}
Cross-view geo-localization is similar to other localization techniques such as GPS which is based on satellite or Enhanced Observed Time Difference (E-OTD) which is localized via the cellular networks that provide localization service for the users. As discussed in the main text, cross-view geo-localization can be applied in many areas such as autonomous driving, unmanned aerial vehicle navigation, and augmented reality. It can also serve as an auxiliary technique for positioning smart devices in metropolitan areas where GPS signals and mobile phone signals are usually weak due to dense buildings. It is also worth mentioning that there could be potential exposure of users' private location information due to the nature of the localization technique. Also, there might be exposure of other information such as license plate, vehicle model, etc. However, with proposer anonymous approaches, our proposed method can benefit society in many aspects.

\section{Implementation Details}

\subsection{Implementation Details}
We implement our proposed model in Pytorch~\citep{pytorch}. The ground images and aerial images are resized to $122 \times 671$ and $256 \times 256$, respectively. Similar to previous methods~\citep{SAFA,DSM,l2ltr,comingDown,cvmnet}, we set the batch size to $32$ and $\alpha$ to $10$ in soft margin triplet loss. Within each batch, the exhaustive mini-batch strategy~\citep{Vo} is utilized for constructing triplet pairs. We use AdamW~\citep{adamw} to train our algorithm for $200$ epochs with a weight decay of $0.03$ on a single Nvidia V100 GPU. The learning rate is chosen to be $10^{-4}$ with the cosine learning rate schedule. The $\beta$ in counterfactual loss is set to $5$. The backbone ResNet34~\citep{resnet} is pretrained on ImageNet~\citep{imagenet}. We adopt a $2$-layer transformer with $4$ heads for the geometric layout extractor which is initialized randomly. The number of geometric descriptors $K$ is set to $8$.

\subsection{Semantic augmentation}
We implement our semantic augmentation by using the TorchVision\footnote{\url{https://pytorch.org/vision/stable/index.html}} library. We use {\it ColorJitter} to randomly adjust brightness, contrast, and saturation. We set each parameter to $0.3$ in this function. {\it RandomGrayscale} and {\it RandomPosterize} are used for randomly applying image grayscale and image posterizing during training with a probability of $0.2$. Finally, we apply Gaussian blur with a random kernel size chosen from $\{1,3,5\}$ and a random sigma chosen within $[0.1, 5]$.

\subsection{Geometric Layout Extractor}
As discussed in the main text, our geometric layout extractor is built on the standard transformer. In our implementation, we adopt the Pytorch official implementation of transformer~\footnote{\url{https://pytorch.org/docs/stable/generated/torch.nn.TransformerEncoder.html}}. For each transformer encoder layer, we set the latent dimension to $168$, the feedforward layer dimension to $2048$, and the drop out the probability to $0.3$. We adopt layer norm and $GeLU$ activation function for each layer. At last, a $HardTanh$ activation function is applied before the output of the produced geometric layout descriptors.

\section{Computational Efficiency}

% \begin{wraptable}{l}{0.48\textwidth}
\begin{table}[!t]
    \label{tab:param}
    \centering
    \small
    \setlength{\tabcolsep}{3.5 mm}
    % \begin{tabular}{c|c|c|c|c|c}
    \begin{tabular}{ccc}
    \toprule
     & L2LTR & Ours \\ \midrule
    \# of Parameters & 195.9M & \textbf{48.5M} \\ \midrule
    Inference Time & 405ms & \textbf{235ms} \\ \midrule
    Pretraining & \makecell{ImageNet-21K \\ for ViT} & \makecell{\textbf{ImageNet-1K for} \\ \textbf{ResNet34 only}} \\ \midrule
    Computational Cost & 46.77 GFLOPS & \textbf{39.89 GFLOPS} \\
    \bottomrule
    \end{tabular}
% \end{wraptable}
\caption{Comparison between \ourmodel{} and L2LTR~\citep{l2ltr} on the number of parameters, inference time, initialization model, and computational cost. All experiments are conducted on a single Nvidia V100 GPU with batch size 32.}
\end{table}

In this section, we directly compare the number of parameters, inference time, and pretrained weight between \ourmodel{} and current state-of-the-art methods L2LTR~\citep{l2ltr}. The number of parameters of \ourmodel{} is only $48.5M$ which is $5\times$ fewer parameters than L2LTR ($195.9M$). The inference time for \ourmodel{} is $235$ ms and L2LTR is $405$ ms. We almost double the inference speed compared with L2LTR. We also measure the computational cost in terms of FLOPS. Our method achieves $39.89$ 
GFLOPS, while L2LTR needs $46.77$ GFLOPS. Finally, \ourmodel{} only needs ResNet34~\citep{resnet} pretrained model for the backbone and is randomly initialized for the other components. L2LTR is initialized with weights from ViT~\citep{ViT} which is pretrained on ImageNet-21K.

\section{More Ablation Study}
In this section, we present the complete results of ablation study on different LS levels, different number of geometric layout descriptors, applying LS to other models, the effectiveness of the counterfactual process, and different transformer architectures.

\subsection{Different LS levels}

\begin{table*}[!t]
\centering
\small
\setlength{\tabcolsep}{2 mm}
\begin{tabular}{cccccccccc}
    \toprule 
    % \multirow{2}{*}{} & \multirow{2}{*}{$N_{des.}$} &
    \multicolumn{2}{l}{\textbf{\aug{} levels}} &
    \multicolumn{4}{c}{Same-area} &
    \multicolumn{4}{c}{Cross-area} \\
    \cmidrule(lr){3-6}
    \cmidrule(lr){7-10}
     & Configuration & \Rone{} & \Rfive{} & \Rten{} & \Ronep & \Rone{} & \Rfive{} & \Rten{} & \Ronep \\
    \midrule
    \multirow{9}{*}{
        \begin{tabular}{@{}c@{}}
            \rotatebox[origin=c]{90}{\makecell{Trained on \\CVUSA}}
        \end{tabular}
    } & (none, none)   & 95.23\% & 98.71\% & 99.26\% & 99.79\% & 47.79\% & 70.52\% & 77.52\% & 92.20\% \\
      & (none, weak)   & 95.42\% & 98.01\% & 99.34\% & 99.80\% & 49.08\% & 72.46\% & 79.31\% & 92.95\% \\
      & (none, strong) & 94.31\% & 98.41\% & 99.16\% & 99.80\% & 51.73\% & 73.75\% & 80.42\% & 93.22\% \\
      & (weak, none)   & 95.90\% & 99.00\% & 99.38\% & 99.84\% & 49.23\% & 72.17\% & 78.91\% & 92.62\% \\
      & (weak, weak)   & 95.86\% & 99.04\% & 99.38\% & 99.84\% & 49.47\% & 72.34\% & 79.22\% & 93.48\% \\
      & (weak, strong) & 95.24\% & 98.76\% & 99.27\% & 99.84\% & 51.25\% & 73.71\% & 80.31\% & 93.57\% \\
      & (strong, none) & 95.90\% & 99.03\% & 99.41\% & 99.84\% & 50.54\% & 74.00\% & 80.71\% & 93.10\% \\
      & (strong, weak) & 96.09\% & 99.11\% & 99.43\% & 99.83\% & 53.19\% & 75.15\% & 81.13\% & 94.00\% \\
      & (strong, strong) & 95.43\% & 98.86\% & 99.34\% & 99.86\% & 53.16\% & 75.62\% & 81.90\% & 93.80\% \\
    \midrule
    \multirow{9}{*}{
        \begin{tabular}{@{}c@{}}
            \rotatebox[origin=c]{90}{\makecell{Trained on \\CVACT}}
        \end{tabular}
    } & (none, none)   & 87.42\% & 95.37\% & 96.50\% & 98.65\% & 29.13\% & 47.86\% & 56.21\% & 81.09\% \\
      & (none, weak)   & 87.72\% & 95.20\% & 96.29\% & 98.41\% & 32.52\% & 53.89\% & 62.88\% & 86.39\% \\
      & (none, strong) & 86.67\% & 95.23\% & 96.35\% & 98.38\% & 42.00\% & 64.41\% & 72.32\% & 91.70\% \\
      & (weak, none)   & 86.86\% & 95.37\% & 96.69\% & 98.85\% & 29.15\% & 47.17\% & 54.78\% & 78.40\% \\
      & (weak, weak)   & 87.37\% & 95.84\% & 96.96\% & 98.72\% & 35.65\% & 55.81\% & 63.86\% & 85.74\% \\
      & (weak, strong) & 87.04\% & 95.70\% & 96.88\% & 98.63\% & 44.63\% & 64.89\% & 72.53\% & 90.62\% \\
      & (strong, none) & 86.75\% & 95.69\% & 96.76\% & 98.81\% & 25.71\% & 42.93\% & 50.86\% & 75.16\% \\
      & (strong, weak) & 86.84\% & 95.71\% & 96.83\% & 98.76\% & 35.31\% & 55.71\% & 64.59\% & 88.01\% \\
      & (strong, strong) & 86.21\% & 95.44\% & 96.72\% & 98.77\% & 44.07\% & 64.66\% & 72.08\% & 90.09\% \\
    \bottomrule
\end{tabular}
\caption{
Performance of \ourmodel{} under different \aug{} levels. 
Configurations are marked by the tuple: $(\mathcal{L}_L, \mathcal{L}_S)$, with $\mathcal{L}_L, \mathcal{L}_S \in \{\text{none}, \text{weak}, \text{strong}\}$.
All the models are trained with PT and CF. 
Results are shown for both cases when our models are trianed on CVUSA and CVACT datasets.
}
\label{tab:ablation_ls_levels}
\end{table*}

The experiment with different layout simulation and semantic augmentation (LS) levels (denoted as $\mathcal{L}_L$ and $\mathcal{L}_S$, respectively) is shown in \Cref{tab:ablation_ls_levels}. In these experiments we define three levels (none, weak, strong) for LS. ``none'' stands for no augmentation applied. We only apply random flip in ``weak'' for layout simulation. Similarly, less intensive color jitter (each parameter is set to $0.1$) and only image grayscale with a probability of $0.1$ are adopted in weak semantic augmentation. ``strong'' stands for applying all augmentation methods we discussed in our main paper.

\subsection{Number of descriptors}

\begin{table*}[!t]
\centering
\small
\setlength{\tabcolsep}{2 mm}
\begin{tabular}{cccccccccc}
    \toprule 
    % \multirow{2}{*}{} & \multirow{2}{*}{$N_{des.}$} &
    \multicolumn{2}{l}{\textbf{\# descriptors}} &
    \multicolumn{4}{c}{Same-area} &
    \multicolumn{4}{c}{Cross-area} \\
    \cmidrule(lr){3-6}
    \cmidrule(lr){7-10}
     & Configuration & \Rone{} & \Rfive{} & \Rten{} & \Ronep & \Rone{} & \Rfive{} & \Rten{} & \Ronep \\
    \midrule
    \multirow{4}{*}{
        \begin{tabular}{@{}c@{}}
            \rotatebox[origin=c]{90}{\makecell{Trained on \\CVUSA}}
        \end{tabular}
    } & $K=1$ & 93.44\% & 98.18\% & 99.05\% & 99.80\% & 39.93\% & 63.54\% & 71.59\% & 88.92\% \\
      & $K=2$ & 94.09\% & 98.44\% & 99.13\% & 99.80\% & 43.07\% & 67.30\% & 74.52\% & 90.32\% \\
      & $K=4$ & 94.80\% & 98.64\% & 99.30\% & 99.82\% & 46.36\% & 69.37\% & 76.46\% & 91.27\% \\
      & $K=8$ & 95.43\% & 98.86\% & 99.34\% & 99.86\% & 53.16\% & 75.62\% & 81.90\% & 93.80\% \\
    \midrule
    \multirow{4}{*}{
        \begin{tabular}{@{}c@{}}
            \rotatebox[origin=c]{90}{\makecell{Trained on \\CVACT}}
        \end{tabular}
    } & $K=1$ & 82.08\% & 94.15\% & 95.87\% & 98.63\% & 28.28\% & 46.84\% & 54.95\% & 78.56\% \\
      & $K=2$ & 85.00\% & 95.05\% & 96.53\% & 98.78\% & 37.72\% & 57.16\% & 65.21\% & 85.37\% \\
      & $K=4$ & 85.85\% & 95.50\% & 96.63\% & 98.71\% & 43.30\% & 63.90\% & 71.34\% & 88.88\% \\
      & $K=8$ & 86.21\% & 95.44\% & 96.72\% & 98.77\% & 44.07\% & 64.66\% & 72.08\% & 90.09\% \\
    \bottomrule
\end{tabular}
\caption{
Performance of \ourmodel{} under different numbers of descriptors $K$. 
Configurations include $K=1,2,4,8$.
All the models are trained with PT, \aug{}, and CF. 
Results are shown for both cases when our models are trianed on CVUSA and CVACT datasets.
}
\label{tab:ablation_descriptors_supp}
\end{table*}
In \Cref{tab:ablation_descriptors_supp}, we show the full experiment results under different numbers of layout descriptors. It is observable that, increasing the number of descriptors, significantly improves the cross-area performance.

\subsection{LS for other models}
In \Cref{tab:ablation_ls_levels}, we apply LS techniques to the existing models~\citep{l2ltr,SAFA}. We observe that LS bring substantial improvement on these models especially on cross-area benchmarks. This reflects the generic applicability of the proposed LS technique.

\begin{table*}[!t]
\centering
\small
\setlength{\tabcolsep}{1.4 mm}
\begin{tabular}{cccccccccc}
    \toprule 
    % \multirow{2}{*}{} & \multirow{2}{*}{Method} & 
    \multicolumn{2}{l}{\textbf{\aug{} $+$ other methods}} &
    \multicolumn{4}{c}{Same-area} &
    \multicolumn{4}{c}{Cross-area} \\
    \cmidrule(lr){3-6}
    \cmidrule(lr){7-10}
     & Configuration & \Rone{} & \Rfive{} & \Rten{} & \Ronep & \Rone{} & \Rfive{} & \Rten{} & \Ronep \\
    \midrule
    \multirow{6}{*}{
        \begin{tabular}{@{}c@{}}
            \rotatebox[origin=c]{90}{\makecell{Trained on \\CVUSA}}
        \end{tabular}
    } 
    & SAFA & 89.84\% & 96.93\% & 98.14\% & 99.64\% & 30.40\% & 52.93\% & 62.29\% & 85.82\% \\
      & SAFA w/ \aug{} & 88.19\% & 96.48\% & 98.20\% & 99.74\% & 37.15\% & 60.31\% & 69.20\% & 89.15\% \\[2mm]
      & L2LTR &	94.05\% & 98.27\% & 98.99\% & 99.67\% & 47.55\% & 70.58\% & 77.52\% & 91.39\% \\
      & L2LTR w/ \aug{} & 93.62\% & 98.46\% & 99.03\% & 99.77\% & 52.58\% & \FT{75.81\%} & \FT{82.19\%} & 93.51\% \\[2mm]
      & \ourmodel{} w/o \aug{} & 95.23\% & 98.71\% & 99.26\% & 99.79\% & 47.79\% & 70.52\% & 77.52\% & 92.20\% \\
      & \ourmodel{} w/ \aug{} & \FT{95.43\%} & \FT{98.86\%} & \FT{99.34\%} &  \FT{99.86\%} & \FT{53.16\%} & 75.62\% & 81.90\% & \FT{93.80\%} \\
    \midrule
    \multirow{6}{*}{
        \begin{tabular}{@{}c@{}}
            \rotatebox[origin=c]{90}{\makecell{Trained on \\CVACT}}
        \end{tabular}
    } 
    & SAFA & 81.03\% & 92.80\% & 94.84\% & 98.17\% & 21.45\% & 36.55\% & 43.79\% & 69.83\% \\
      & SAFA w/ \aug{} & 79.88\% & 92.84\% & 94.71\% & 97.96\% & 25.42\% & 42.30\% & 50.36\% & 76.49\% \\[2mm]
      & L2LTR & 84.89\% & 94.59\% & 95.96\% & 98.37\% & 33.00\% & 51.87\% & 60.63\% & 84.79\% \\
      & L2LTR w/ \aug{} & 83.49\% & 94.93\% & 96.44\% & 98.68\% & 37.69\% & 57.78\% & 66.22\% & 89.63\% \\[2mm]
      & \ourmodel{} w/o \aug{} & \FT{87.42\%} & 95.37\% & 96.50\% & 98.65\% & 29.13\% & 47.86\% & 56.21\% & 81.09\% \\
      & \ourmodel{} w/ \aug{} & 86.21\% & \FT{95.44\%} & \FT{96.72\%} & \FT{98.77\%} & \FT{44.07\%} & \FT{64.66\%} & \FT{72.08\%} & \FT{90.09\%} \\
    \bottomrule
\end{tabular}
\caption{
Comparison of performance of SAFA~\citep{SAFA}, L2LTR~\citep{l2ltr} and \ourmodel{} that are trained with or without the proposed \aug{} technique. 
Our models are trained with PT and CF.
Results are shown for both cases when the models are trained on CVUSA and CVACT datasets. 
% \FT{Best} results are shown in bold.
}
\label{tab:ablation_ls_models}
\end{table*}

\subsection{Counterfactual loss}
The effectiveness of the proposed counterfactual (CF) learning process is demonstrated in~\Cref{tab:ablation_cf_loss}. We conduct an ablation study of CF on models training with and without LS. We observe that, in either case, CF boosts the performance of our model on both same-area and cross-area benchmarks. 

\begin{table*}[!t]
\centering
\small
\setlength{\tabcolsep}{2 mm}
\begin{tabular}{cccccccccc}
    \toprule 
    % \multirow{2}{*}{} & \multirow{2}{*}{$N_{des.}$} &
    \multicolumn{2}{l}{\textbf{CF loss}} &
    \multicolumn{4}{c}{Same-area} &
    \multicolumn{4}{c}{Cross-area} \\
    \cmidrule(lr){3-6}
    \cmidrule(lr){7-10}
     & Configuration & \Rone{} & \Rfive{} & \Rten{} & \Ronep & \Rone{} & \Rfive{} & \Rten{} & \Ronep \\
    \midrule
    \multirow{4}{*}{
        \begin{tabular}{@{}c@{}}
            \rotatebox[origin=c]{90}{\makecell{Trained on \\CVUSA}}
        \end{tabular}
    } 
      & (CF, w/ \aug{}) & 95.43\% & 98.86\% & 99.34\% & 99.86\% & 53.16\% & 75.62\% & 81.90\% & 93.80\% \\
      & (noCF, w/ \aug{}) & 95.06\% & 98.72\% & 99.30\% & 99.85\% & 49.18\% & 71.96\% & 79.28\% & 92.84\% \\
      & (noCF, w/o \aug{}) & 94.83\% & 98.59\% & 99.28\% & 99.80\% & 43.71\% & 66.87\% & 74.68\% & 90.66\% \\
      & (CF, w/o \aug{}) & 95.23\% & 98.71\% & 99.26\% & 99.79\% & 47.79\% & 70.52\% & 77.52\% & 92.20\% \\
    \midrule
    \multirow{4}{*}{
        \begin{tabular}{@{}c@{}}
            \rotatebox[origin=c]{90}{\makecell{Trained on \\CVACT}}
        \end{tabular}
    } 
      & (8, CF, w/ \aug{}) & 86.21\% & 95.44\% & 96.72\% & 98.77\% & 44.07\% & 64.66\% & 72.08\% & 90.09\% \\
      & (8, noCF, w/ \aug{}) & 85.84\% & 95.48\% & 96.67\% & 98.71\% & 43.61\% & 64.00\% & 71.57\% & 90.04\% \\
      & (8, noCF, w/o \aug{}) & 87.01\% & 95.05\% & 96.45\% & 98.40\% & 28.26\% & 46.92\% & 55.26\% & 81.21\% \\
      & (8, CF, w/o \aug{}) & 87.42\% & 95.37\% & 96.50\% & 98.65\% & 29.13\% & 47.86\% & 56.21\% & 81.09\% \\
    \bottomrule
\end{tabular}
\caption{
Performance of \ourmodel{} under different configurations, including
with or without counterfactual loss (CF or noCF), and with or without \aug{} (w/ \aug{} or w/o \aug{}).
% on same-area and cross-area benchmarks. 
All the models are trained with PT.
Results are shown for both cases when our model is trianed on CVUSA and CVACT datasets.
}
\label{tab:ablation_cf_loss}
\end{table*}

\subsection{Transformer architecture}
Finally, we vary the architecture of the transformer in our geometric layout extractor in~\Cref{tab:ablation_tr_arch}. Noticeably, we observe that with increasing the number of transformer heads, the performance of our model is improved as well. Similarly, by using transformer layers it also boosts the performance. With the configuration of $4$ heads and $2$ layers, our model almost converges to a stable performance.

\begin{table*}[!t]
\centering
\small
\setlength{\tabcolsep}{2.0 mm}
\begin{tabular}{cccccccccc}
    \toprule 
    \multicolumn{2}{l}{\textbf{TR ARCH.}} &
    \multicolumn{4}{c}{Same-area} &
    \multicolumn{4}{c}{Cross-area} \\
    \cmidrule(lr){3-6}
    \cmidrule(lr){7-10}
     & Configuration & \Rone{} & \Rfive{} & \Rten{} & \Ronep & \Rone{} & \Rfive{} & \Rten{} & \Ronep \\
    \midrule
    \multirow{6}{*}{
        \begin{tabular}{@{}c@{}}
            \rotatebox[origin=c]{90}{\makecell{Trained on \\CVUSA}}
        \end{tabular}
    } & $(8, 0, 0)$  & 92.67 \% & 98.19 \% & 99.04 \% & 99.78 \% & 46.88 \% & 70.81 \% & 78.36 \% & 92.99\\
      & $(1, 1, 1)$ & 93.12\% & 98.19\% & 98.89\% & 99.80\% & 34.30\% & 58.89\% & 67.26\% & 87.39\% \\
      & $(1, 2, 1)$ & 93.48\% & 98.36\% & 99.09\% & 99.81\% & 41.16\% & 64.73\% & 73.18\% & 90.00\% \\
      & $(1, 4, 1)$ & 93.43\% & 98.28\% & 99.04\% & 99.76\% & 38.60\% & 62.48\% & 70.72\% & 88.90\% \\
      & $(1, 4, 2)$ & 93.44\% & 98.18\% & 99.05\% & 99.80\% & 39.93\% & 63.54\% & 71.59\% & 88.92\% \\
      & $(8, 4, 2)$ & 95.43\% & 98.86\% & 99.34\% & 99.86\% & 53.16\% & 75.62\% & 81.90\% & 93.80\% \\
    \midrule
    \multirow{6}{*}{
        \begin{tabular}{@{}c@{}}
            \rotatebox[origin=c]{90}{\makecell{Trained on \\CVACT}}
        \end{tabular}
    } & $(8, 0, 0)$ & 83.96 \% & 94.29 \% & 96.11 \% & 98.60 \% & 37.66 \% & 59.57 \% & 67.86 \% & 89.69 \\
      & $(1, 1, 1)$ & 82.70\% & 94.50\% & 95.89\% & 98.57\% & 24.83\% & 44.63\% & 53.22\% & 80.17\% \\
      & $(1, 2, 1)$ & 83.13\% & 94.47\% & 96.22\% & 98.71\% & 25.62\% & 44.05\% & 52.78\% & 78.20\% \\
      & $(1, 4, 1)$ & 82.68\% & 94.42\% & 95.90\% & 98.63\% & 25.20\% & 43.47\% & 52.12\% & 77.27\% \\
      & $(1, 4, 2)$ & 82.08\% & 94.15\% & 95.87\% & 98.63\% & 28.28\% & 46.84\% & 54.95\% & 78.56\% \\
      & $(8, 4, 2)$ & 86.21\% & 95.44\% & 96.72\% & 98.77\% & 44.07\% & 64.66\% & 72.08\% & 90.09\% \\
    \bottomrule
\end{tabular}
\caption{
Performance of \ourmodel{} under different settings of the geometric layout extractor module $\mathcal{G}^v(\cdot)$. 
Configurations are marked by the tuple: $(K, N_h, N_l)$ where $K$ referes to the
the number of descriptors, $N_h$ the number of heads in the transformer encoder layer, and $N_l$ the number of transformer encoder layers, respectively.
% $(8,0,0)$ indicates the case without any transformer encoder layer.
% on same-area and cross-area benchmarks. 
All the models are trained with PT, \aug{}, and CF.
Results are shown for both cases when our model is trianed on CVUSA and CVACT datasets.}
\label{tab:ablation_tr_arch}
\end{table*}

\subsection{Smaller latent feature dimensions}

We have conducted an ablation study with a $1024$ latent feature dimension which is $4$ times less than the model we used in the main script. To achieve this goal, we replace the last convolution layer of the backbone network which outputs $512$ channels with a randomly initialized convolution layer which outputs $128$ channels before training. The same-area and cross-area performance on both CVUSA and CVACT datasets are shown in \Cref{tab:small_feat}. As shown from the results, the proposed method with a smaller latent feature dimension, $\text{\ourmodel{}}_s$, retains the performance on the same-area experiment and only shows a small performance drop ~$2\%$ on the cross-area experiment. More importantly, our model with smaller latent feature dimensions still achieves state-of-the-art performance.

\begin{table*}[!t]
\centering
\small
\setlength{\tabcolsep}{1.4 mm}
\begin{tabular}{ccccccccccc}
    \toprule 
    % \multirow{2}{*}{} & \multirow{2}{*}{Method} & 
    \multicolumn{3}{l}{\textbf{Smaller feature dimension}} &
    \multicolumn{4}{c}{Same-area} &
    \multicolumn{4}{c}{Cross-area} \\
    \cmidrule(lr){4-7}
    \cmidrule(lr){8-11}
     & Method & Dim & \Rone{} & \Rfive{} & \Rten{} & \Ronep & \Rone{} & \Rfive{} & \Rten{} & \Ronep \\
    \midrule
    \multirow{6}{*}{
        \begin{tabular}{@{}c@{}}
            \rotatebox[origin=c]{90}{\makecell{Trained on \\CVUSA}}
        \end{tabular}
    } 
    & SAFA & 4096 & 89.84\% & 96.93\% & 98.14\% & 99.64\% & 30.40\% & 52.93\% & 62.29\% & 85.82\% \\
    & L2LTR & 768 &	94.05\% & 98.27\% & 98.99\% & 99.67\% & 47.55\% & 70.58\% & 77.52\% & 91.39\% \\
    & TransGeo & 1024 & 94.08\% & 98.36\% &	99.04\% & 99.77\% & 37.18\% & 61.57\% & 69.86\% & 89.14\% \\
    & $\text{\ourmodel{}}_s$ & 1024 & \SB{94.61\%} & \SB{98.57\%} & \SB{99.21\%} & \SB{99.83\%} & \SB{50.31\%} & \SB{73.34\%} & \SB{79.38\%} & \SB{92.96\%} \\
    & \ourmodel{} & 4096 & \B{95.43\%} & \B{98.86\%} & \B{99.34\%} &  \B{99.86\%} & \B{53.16\%} & \B{75.62\%} & \B{81.90\%} & \B{93.80\%} \\
    \midrule
    \multirow{6}{*}{
        \begin{tabular}{@{}c@{}}
            \rotatebox[origin=c]{90}{\makecell{Trained on \\CVACT}}
        \end{tabular}
    } 
    & SAFA & 4096 & 81.03\% & 92.80\% & 94.84\% & 98.17\% & 21.45\% & 36.55\% & 43.79\% & 69.83\% \\
    & L2LTR & 768 & 84.89\% & 94.59\% & 95.96\% & 98.37\% & 33.00\% & 51.87\% & 60.63\% & 84.79\% \\
    & TransGeo & 1024 & 84.95\% & 94.14\% & 95.78\% & 98.37\% & 17.45\% & 32.49\% & 40.48\% & 69.14\% \\
    & $\text{\ourmodel{}}_s$ & 1024 & \B{86.24\%} & \B{95.62\%} & \B{96.75\%} & \SB{98.59\%} & \SB{42.32\%} & \SB{62.24\%} & \SB{70.17\%} & \SB{88.62\%} \\
    & \ourmodel{} & 4096 & \SB{86.21\%} & \SB{95.44\%} & \SB{96.72\%} & \B{98.77\%} & \B{44.07\%} & \B{64.66\%} & \B{72.08\%} & \B{90.09\%} \\
    \bottomrule
\end{tabular}
\caption{
Comparison of our model with different latent feature dimensions with SAFA~\citep{SAFA}, L2LTR~\citep{l2ltr}, and TransGeo~\citep{transgeo}. $\text{\ourmodel{}}_s$ represents the proposed model with smaller latent feature dimension. The \B{best} results are shown in magenta and the \SB{second best} results are shown in blue.
}
\label{tab:small_feat}
\end{table*}

\section{More Qualitative Results}
\begin{figure*}[!t]
    \centering
    \includegraphics[width=0.9\textwidth, page=2, trim={0cm 0cm 8cm 0cm}]{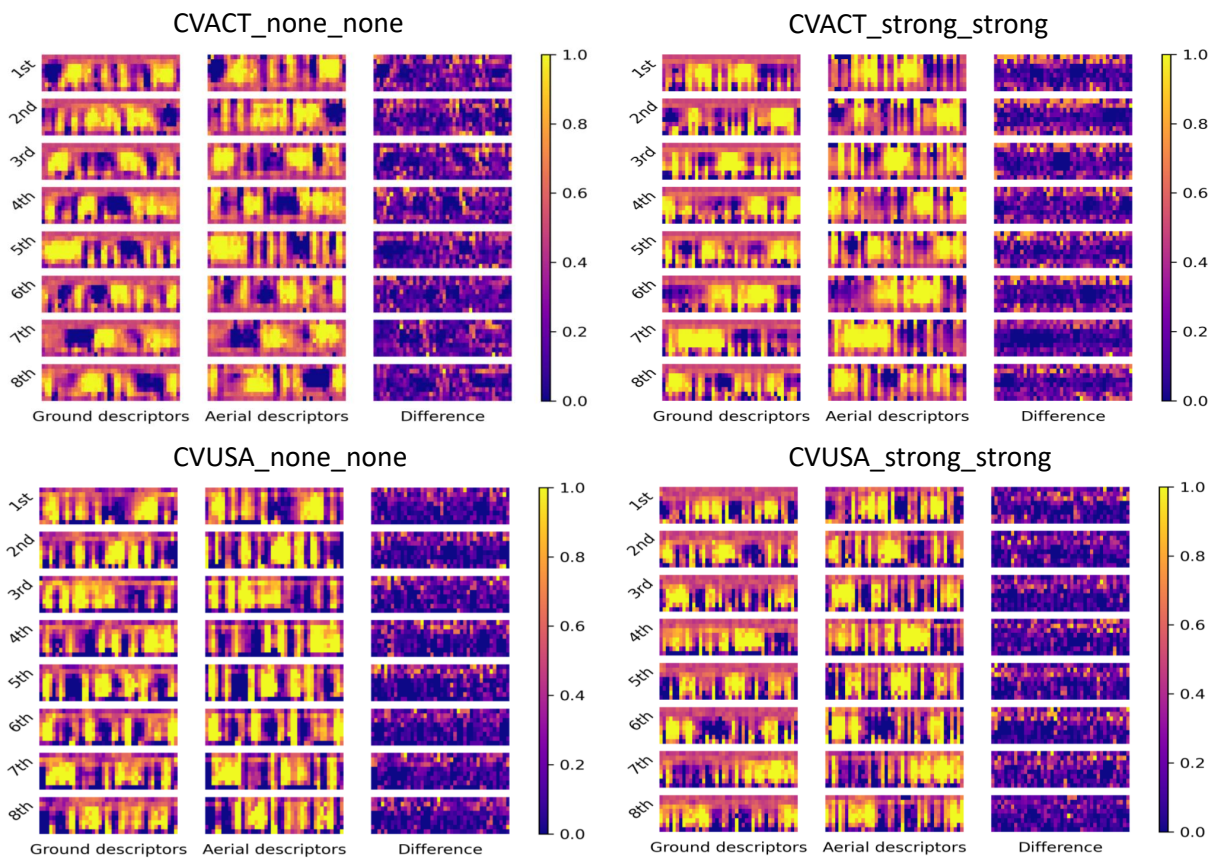}
    \caption{More visualization of geometric layout descriptors from our model trained with polar transformation on CVUSA and CVACT dataset the title stands for \textit{dataset\_layout simulation\_semantic augmentation}.}
    \label{fig:more_polar}
\end{figure*}

\begin{figure*}[!ht]
    \centering
    \includegraphics[width=0.9\textwidth, page=3, trim={1cm 0cm 7.5cm 0cm}]{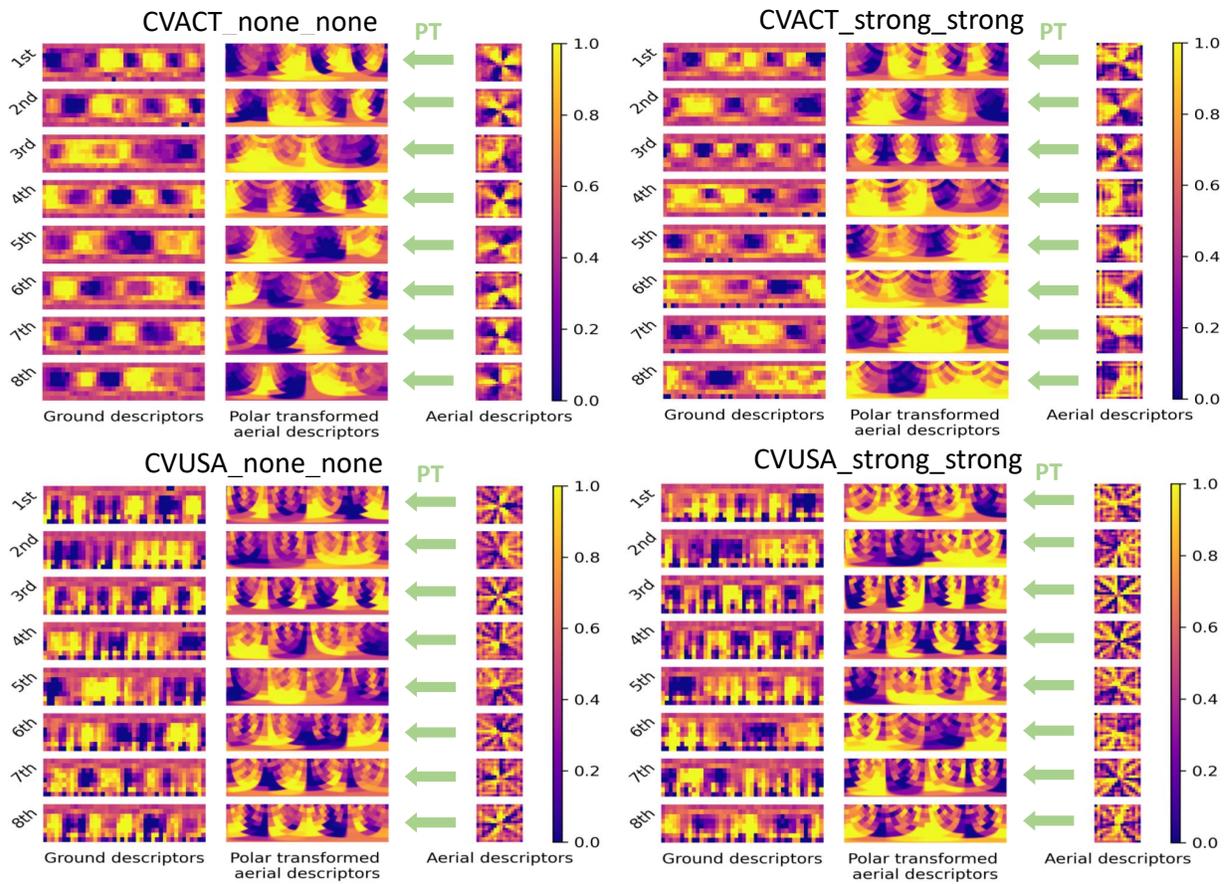}
    \caption{More visualization of geometric layout descriptors from our model trained without polar transformation on CVUSA and CVACT dataset. The naming convention is the same as in~\Cref{fig:more_polar}.}
    \label{fig:more_nopolar}
\end{figure*}

More qualitative visualization of learned geometric layout descriptors are presented in~\Cref{fig:more_polar,fig:more_nopolar}. We present models trained on different dataset and varying LS configurations. \Cref{fig:more_polar} shows descriptors from models trained with polar transformation and \Cref{fig:more_nopolar} shows descriptors from models trained without polar transformation. We can observe that the correspondence holds in different configurations of our \ourmodel{} on different training data.

\end{document}